\documentclass[10pt,twocolumn,letterpaper]{article}

\usepackage[pagenumbers]{cvpr}

\usepackage{cuted}

\usepackage{caption}

\makeatletter\AtBeginDocument{\renewcommand\paragraph{\@startsection{paragraph}{4}{\z@}{1ex \@plus.3ex \@minus.2ex}{-1em}%
  {\normalfont\normalsize\bfseries}}}\makeatother

\newcommand{\method}{DynMF2\xspace}
\usepackage{pifont}
\usepackage{array}
\newcommand{\cmark}{\ding{51}}
\newcommand{\xmark}{\ding{55}}

\DeclareMathOperator*{\argmin}{arg\,min}

\usepackage[protrusion=true,expansion=true]{microtype}
\AtBeginDocument{\microtypesetup{protrusion=false,expansion=false}}
\definecolor{cvprblue}{rgb}{0.21,0.49,0.74}
\usepackage[pagebackref,breaklinks,colorlinks,allcolors=cvprblue]{hyperref}

\def\confName{CVPR}
\def\confYear{2027}

\title{\vspace*{-11pt}\method: Rigid Motion Factorization by Expectation-Maximization \\ for Monocular 4D Reconstruction and Motion Segmentation}

\author{Agelos Kratimenos$^{1*}$ \qquad Yile Hua$^{1*}$ \qquad Kostas Daniilidis$^{1,2}$\\[4pt]
{\normalsize $^{1}$University of Pennsylvania \qquad $^{2}$Robotics Institute, Athena RC}\\[-20pt]
}

\begin{document}
\maketitle
\begin{strip}
\centering
\setlength{\tabcolsep}{0pt}%
\begin{tabular}{@{}c@{\hspace{6pt}}c@{\hspace{6pt}}c@{}}
\makebox[1.64in]{\includegraphics[height=1.95in]{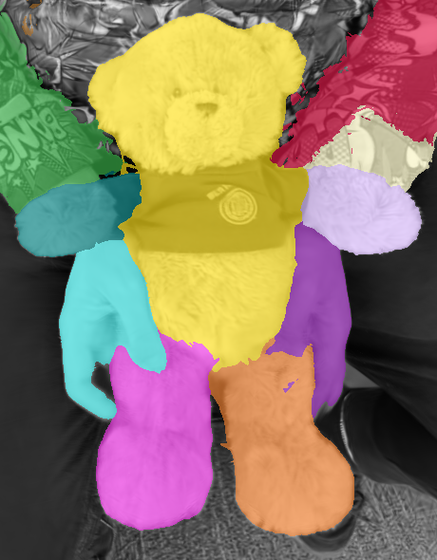}} & \includegraphics[height=1.95in]{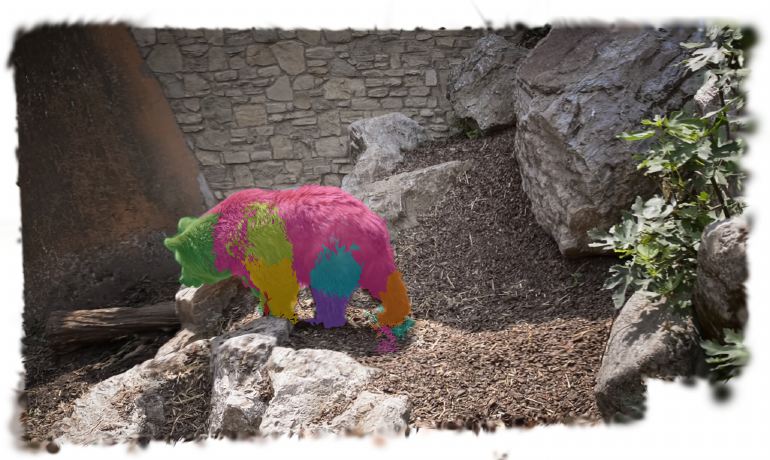} &
\makebox[1.64in]{\includegraphics[height=1.95in]{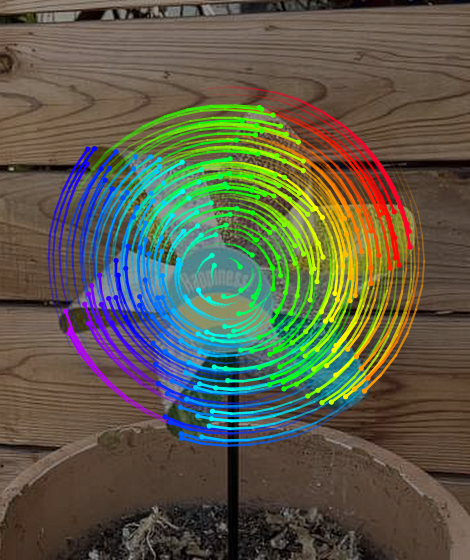}} \\[1pt]
\makebox[1.64in]{\small (a) $\mathrm{SE}(3)$ motion segmentation} & \small (b) 4D reconstruction & \makebox[1.64in]{\small (c) 3D tracking}
\end{tabular}
\captionsetup{hypcap=false}\setlength{\abovecaptionskip}{4pt}%
\captionof{figure}{\textbf{\method models a single casual video with a handful of rigid motions, each Gaussian following exactly
one, without masks.} (a)~Our motions on \emph{teddy}, one colour each. (b)~Our 4D reconstruction of
\emph{bear}, coloured by motion. (c)~3D tracks on \emph{paper-windmill}.}
\captionsetup{hypcap=true}
\label{fig:teaser}%
\vspace{-1mm}%
\end{strip}
{\renewcommand{\thefootnote}{*}\footnotetext{Equal contribution.}}
\begin{abstract}
A single casual video leaves a moving scene under-determined: a point seen once from each place can slide toward or away
from the camera and still match every frame. We constrain the reconstruction with the structure of the motion itself.
Building on DynMF, we model a scene's motion as a handful of $\mathrm{SE}(3)$ trajectories, each rigid, and let every point follow
one of them alone, never a blend. \method computes this factorization directly from 3D point trajectories, with no
rendering, by Expectation-Maximization (EM). Its M-step fits, in closed form, every rigid motion to the points assigned to
it, and its E-step reassigns every point to the motion that explains it with the least error. The background is simply one
more motion, the camera's, so no masks are needed. Three tools then revise the factorization: requery adds trajectories
where the rendering shows parts the tracker missed, merge joins motions that move together, and split separates those
that do not. Thus the number of motions follows the scene. Without masks, \method clearly outperforms the mask-based baselines in view synthesis on iPhone and NVIDIA and in 3D tracking on iPhone. It recovers what masks leave out, from thin parts
and details to small movers far in the background. Its motions segment the moving parts of MOVi-F and iTACO scenes
competitively, although \method is never trained to segment. What moves in a scene is not decided in advance: it
emerges from the rigid motions themselves.
\end{abstract}
\vspace{-4mm}%
\section{Introduction}
\label{sec:intro}

Recovering a moving scene from one casual video, to see it from new viewpoints and follow every point through time,
would serve immersive video~\cite{broxton2020immersive}, dynamic scene editing~\cite{huang2024scgs} and robots that
learn by watching people~\cite{allshire2025videomimic}. Deformation fields carried radiance fields and 3D Gaussians from
static scenes to moving
ones~\cite{mildenhall2020nerf,kerbl20233dgs,pumarola2021dnerf,park2021nerfies,yang2024deformable3dgs,wu2024fourdgs},
but on hand-held captures they lose 1 to 2\,dB, and 4 to 5\,dB on complex motion~\cite{gao2022dycheck}. Progress since
has come mostly from priors: per-frame depth~\cite{piccinelli2024unidepth,yang2024depthanything} and point tracks
across the whole video, in 2D~\cite{doersch2024bootstap,karaev2025cotracker3} and now in
3D~\cite{xiao2024spatialtracker,ngo2024delta}. Lifted with depth, the tracks give the motion a starting point that
photometric losses cannot find alone~\cite{wang2024shapeofmotion,lei2024mosca,zhou2026motionscale}. The priors have
improved faster than the way methods use them.

Casual video leaves this reconstruction under-determined. A single camera sees each point from one place at each
instant, so a point that drifts along its line of sight, or fades where it should be hidden, still matches every
training frame, and the error shows only from a new viewpoint. The motion model has to supply the constraint that the
views do not. Many models impose almost none: feed-forward networks predict a point map per
frame~\cite{zhang2025monst3r,wang2025cut3r}, and deformation fields give each Gaussian its own
motion~\cite{yang2024deformable3dgs,wu2024fourdgs}. Others let every Gaussian blend a few shared motions, as global
bases~\cite{kratimenos2023dynmf,wang2024shapeofmotion}, scaffold nodes~\cite{lei2024mosca} or local bases in a
cluster~\cite{zhou2026motionscale}. Shape of Motion~\cite{wang2024shapeofmotion} argues that 3D scene motion
decomposes into a few rigid ones, yet a blend of rigid motions is not rigid. Each Gaussian moves on its own again, and a wrong motion can still fit the training views.

Most methods must also be told what moves. Shape of Motion and MotionScale separate moving objects from the static
background with masks, drawn from user clicks~\cite{wang2024shapeofmotion,yang2023trackanything} or by
SAM2~\cite{zhou2026motionscale,ravi2024sam2}, and MoSca thresholds the epipolar error of optical
flow~\cite{lei2024mosca}. Whatever moves outside the masks or below the threshold, such as a person walking far in the
background, is reconstructed as static. The motions then come from a common recipe: lifted tracks are grouped once by
k-means on their positions or velocities, and Procrustes alignment fits each group with a rigid
motion~\cite{wang2024shapeofmotion,zhou2026motionscale}. The grouping never asks whether a rigid motion explains it,
and the fitted motions never send a track back to a better group. Training only partly repairs this first pass. Parts the tracker missed, often thin or fast ones, never get a motion of their own and borrow a
neighbour's. They refine the motions of a grouping, not the grouping.

Kratimenos \etal~\cite{kratimenos2023dynmf} showed that the motion of a scene factorizes into a few shared trajectories,
but learned those trajectories and each point's weights from scratch, through the renderer. With today's 3D tracks, both can be
computed before training starts.
We argue that the motion of most scenes is a few rigid $\mathrm{SE}(3)$ motions, and that a model should hold every
point to exactly one of them, never a blend. The restriction is deliberate: a point that cannot blend motions cannot
hide a wrong one either. We keep the common start, k-means and Procrustes, and take it one step further, into an EM that alternates
between fitting each motion to its tracks in closed form and moving each track to the motion that predicts it best,
until neither changes. The motions thus choose the grouping, and no mask is needed, since the static background is found
as one more motion. Three tools then keep revising it. Where the rendering disagrees with the video, \method requeries tracks
and gives them motions. It merges motions that move as one, static or not, and splits those that do not, so the number
of motions follows the scene.
Motion segmentation and 3D tracking come out of the model as by-products.

We test \method on two benchmarks of monocular dynamic view synthesis, iPhone~\cite{gao2022dycheck} and
NVIDIA~\cite{yoon2020novel}. Without masks, it renders new views at 18.05 and 23.83\,dB masked PSNR, against
16.72 and 23.37\,dB for Shape of Motion~\cite{wang2024shapeofmotion}, and tracks points in 3D with an
end-point error of 0.058\,m, against 0.082\,m. On MOVi-F~\cite{greff2022kubric}, without segmentation labels, its
motions segment scenes at 0.775 ARI, against 0.579 for the supervised Segment Any Motion~\cite{huang2025segmentanymotion}.
It keeps small movers, such as people walking far behind the subject (\cref{fig:qual}), that methods built on masks
or thresholds freeze into the background. Our contributions are:
\begin{itemize}
\item \method, a monocular 4D reconstruction method that holds each Gaussian to exactly one of a few rigid
$\mathrm{SE}(3)$ motions, with no masks and no separate static set.
\item To our knowledge the first dynamic Gaussian-splatting method whose motions are computed by EM rigid-motion segmentation
of 3D tracks: tracks are grouped by the rigid
motions that predict them, not by k-means alone, and requery, merge and split revise the groups, so that their
number follows the scene.
\item Extensive experiments and ablations on view synthesis, 3D tracking and motion segmentation, showing that hard
rigid motions found without masks compete with or beat methods that blend motions and are told what moves.
\end{itemize}
\section{Related work}
\label{sec:related}
\begin{figure*}[t]
  \centering
  \includegraphics[width=\textwidth]{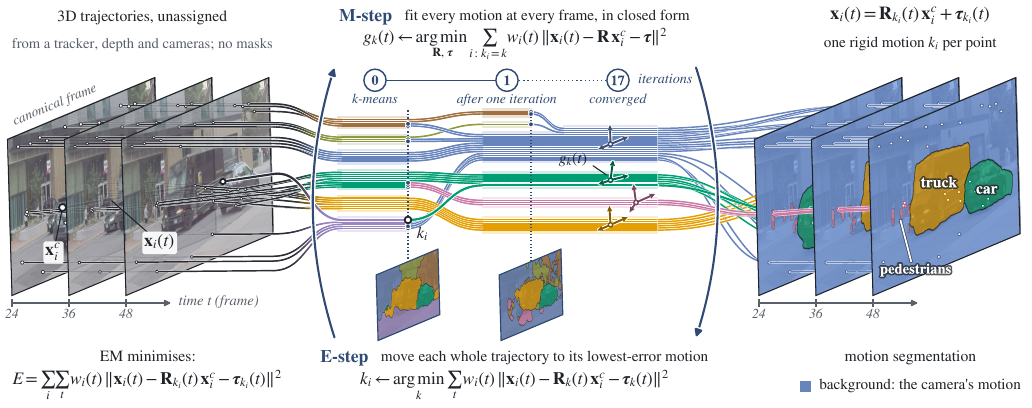}
  \caption{Overview of \method on Truck. \emph{Left:} 3D trajectories from a tracker, depth and the cameras, without masks and
  with no motion yet. \emph{Middle:} EM minimises the energy of \cref{eq:energy}, giving each whole trajectory one rigid
  motion. Left to right are its iterations, each over all frames. From k-means, which splits the background, the M-step
  fits every motion at every frame with \cref{eq:mstep} and the E-step moves each trajectory to its lowest-error motion with
  \cref{eq:estep}, until no $k_i$ changes. \emph{Right:} the trajectories with their motions over the trained model's rendered
  segmentation. The background is the camera's motion.}
  \label{fig:overview}
\vspace{-2mm}%
\end{figure*}

\paragraph{Dynamic view synthesis.}
Radiance fields first reached moving scenes by warping a canonical
space~\cite{pumarola2021dnerf,park2021nerfies,park2021hypernerf} or by linking frames with scene
flow~\cite{li2021nsff,gao2021dynnerf,li2023dynibar}. 3D Gaussians made the scene explicit, moved by a deformation
network or by a trajectory per Gaussian~\cite{luiten2024dynamic3dgs,wu2024fourdgs,yang2024deformable3dgs,stearns2024marbles,wu2025fourdfly,park2026webspline,liang2026reflow}.
Recent work leans on learned priors: depth, cameras and point tracks first came from separate
networks~\cite{piccinelli2024unidepth,li2025megasam,doersch2024bootstap,karaev2025cotracker3,harley2025alltracker,xiao2024spatialtracker,ngo2024delta}.
Single networks now recover cameras and geometry
together~\cite{zhang2025monst3r,wang2025cut3r,wang2025vggt,zhuo2026streamvggt,hu2026vggt4d,wang2026vggtomega}, some
also following points through time~\cite{feng2025st4rtrack,zhang2026d4rt}. Other
networks output renderable Gaussians directly~\cite{liang2025btimer,lin2026movies,xu2025fourdgt,wu2026streamsplat}, while generative
methods fill unseen views with diffusion~\cite{wu2025cat4d,nazarczuk2025vidar,rozumny2026bulletgen,zhu2026worldfrommotion,chen2025cognvs,yu2025trajectorycrafter}.
None of them assigns each point to one rigid motion from a small set.

\paragraph{Structured motion models.}
DynMF~\cite{kratimenos2023dynmf} blends a few shared trajectories with per-Gaussian weights learned through the
renderer. A sparsity loss pushes each Gaussian towards one trajectory. Most later methods keep this mixture, with $\mathrm{SE}(3)$
bases~\cite{wang2024shapeofmotion,wu2026rigs,zhou2025motion4d}, hierarchies~\cite{liang2025himor,li2026gaussiansequences,wang2026worldtree}
or nearby nodes~\cite{huang2024scgs,lei2024mosca,zhang2025motionblender,zhao2026ground4d,yu2026smg,wu2025origs}, even when new bases appear where the render
error is high~\cite{zhang2026se3bsplinegs}, so no point commits to one motion and rigidity is only encouraged.
Others add a motion prior as a loss: a Gaussian process fitted to well-observed
Gaussians~\cite{kim2026gp4dgs}, a graph in which low-uncertainty Gaussians guide the
rest~\cite{guo2026usplat4d}, or one rigid transform per object that a 2D segmenter
labels rigid~\cite{lee2026mogaf}.
SP-GS~\cite{wan2024spgs} does commit each Gaussian to one rigid superpoint, but superpoints are many spatial patches,
driven by a deformation network to speed up rendering. MotionScale~\cite{zhou2026motionscale} commits from positions and
can then only cut: k-means gives hard clusters with soft local bases inside, which training splits and culls.
Articulat3D~\cite{guo2026articulat3d} commits each Gaussian to one part after a soft stage and reassigns it by
gradient, but its parts are the joints of one articulated object.
Grouped 4DGS~\cite{kang2025groupedfourdgs} shares one transform within each group of splats and, from multi-view video,
splits groups and adds splats where render errors cluster. To our
knowledge, none merges motions or queries new tracks during training. \method instead computes the one-hot
weights and the rigid motions by EM on the 3D tracks before training starts, and training revises them in both
directions, with no masks or separate static set. Concurrently, RayCo4D~\cite{ni2026rayco4d} and XM4D~\cite{han2026xm4d} also blend nodes,
the latter computing their motions before rendering. MoSE3~\cite{cheng2026mose3}, also concurrent, brings
$\mathrm{SE}(3)$ into point tracking: trained on synthetic data, it predicts 3D tracks, soft rigidity groups and per-pixel rigid transforms,
but reconstructs no scene.

\paragraph{Motion factorization and segmentation.}
The factorization behind these models is far older. Under an orthographic camera, the tracks of a rigid scene form a
low-rank matrix~\cite{tomasi1992shape}, and bases over shape or trajectory extend this to non-rigid
objects~\cite{bregler2000recovering,akhter2011trajectory,valmadre2012general,zhu2015convolutional}. With several rigid
motions, each one's tracks span their own subspace~\cite{costeira1998multibody,kumar2017spatio}, so finding the motions
becomes grouping the tracks. Subspace clustering fits these subspaces
algebraically~\cite{vidal2005gpca} or writes each track as a sparse combination of the
others~\cite{elhamifar2013sparse}. Spectral clustering of long tracks follows the same
pattern~\cite{brox2010object,ochs2014segmentation}. These methods return a partition, and the motions must be fitted
afterwards. Layered models instead alternate between assigning points to motions and refitting
them~\cite{wang1994layers,ayer1995layered,weiss1996unified,torr1998geometric}, adding or removing models through
energies~\cite{isack2012pearl,delong2012labelcosts,russell2011energy}, which yields both together. Our EM belongs to this
line, as does multi-body SLAM, which clusters rigid bodies in 3D while estimating their
motion~\cite{huang2019clusterslam,huang2020clustervo}, and so does the training-free graph of MotionBits~\cite{qian2026motionbits}. Learned models now segment moving objects from optical
flow~\cite{yang2021motiongrouping,xie2022oclr,lian2023rcf}, from long tracks~\cite{huang2025segmentanymotion} or with
SAM~\cite{xie2024flowsam}, and some carry these classical ideas into their losses, as EM~\cite{meunier2022emdriven} or as
low-rank grouping of trajectories~\cite{karazija2024trajectories}. Other methods also build the scene from the rigid pieces they find: from
monocular tracks~\cite{russell2014videopopup}, with EM over RGB-D frames and object detections~\cite{strecke2019emfusion}, as
primitives cut by masks~\cite{mazur2026primitivemache}, or as parts learned inside a radiance field~\cite{yang2024movingparts}.
\method needs no masks or detections, and it computes its $\mathrm{SE}(3)$ segmentation from tracks instead of learning it
through the renderer; that segmentation is the motion model of the rendered scene.
\section{Method}
\label{sec:method}

\begin{figure*}[t]
  \centering
  \includegraphics[width=\textwidth]{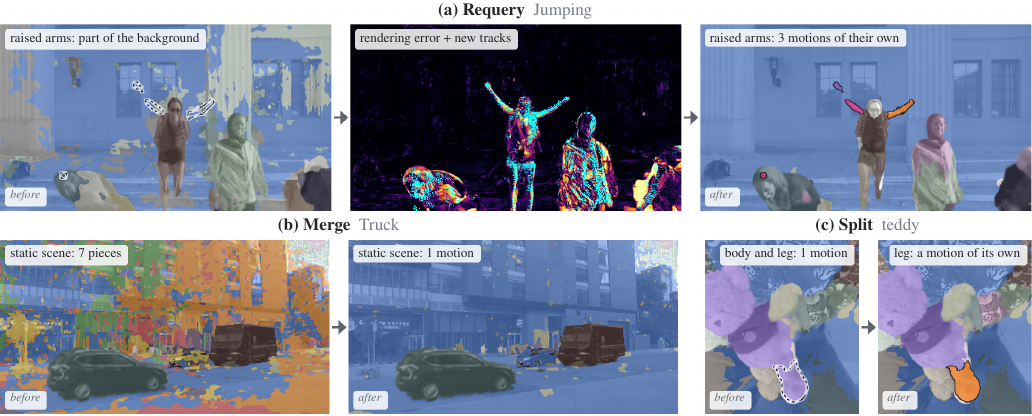}
  \caption{The three tools on real scenes, shown as rendered motion segmentations with one colour per motion. (a)~Requery on
  Jumping, from left to right: the raised arms belong to the background at first, the rendering error marks them and the new
  tracks (cyan) land on them, and after the revision the arms get motions of their own. (b)~Merge on Truck, before and after: seven
  pieces of the background become one motion, and the vehicles keep theirs. (c)~Split on teddy, before and after: the right leg
  leaves the motion of the body.}
  \label{fig:tools}
\vspace{-2mm}%
\end{figure*}

Given the 3D trajectories of points in a video, we explain them by a few rigid motions. DynMF~\cite{kratimenos2023dynmf} writes the trajectory of a point as its canonical position plus a weighted sum
of shared basis trajectories,
\begingroup\setlength{\abovedisplayskip}{4pt plus 1pt minus 1pt}\setlength{\belowdisplayskip}{4pt plus 1pt minus 1pt}%
\begin{equation}
  \mathbf{x}_i(t) = \mathbf{x}_i^c + \sum_j c_{ij}\,\mathbf{b}_j(t),
  \label{eq:dynmf}
\end{equation}\endgroup
and learns every term of it from the images. We make two changes. Each basis becomes a rigid motion
$g_k(t) = (\mathbf{R}_k(t), \boldsymbol{\tau}_k(t)) \in \mathrm{SE}(3)$, and each point follows exactly one of $K$ motions,
given by its assignment $k_i$:
\begingroup\setlength{\abovedisplayskip}{4pt plus 1pt minus 1pt}\setlength{\belowdisplayskip}{4pt plus 1pt minus 1pt}%
\begin{equation}
  \mathbf{x}_i(t) = \mathbf{R}_{k_i}(t)\,\mathbf{x}_i^c + \boldsymbol{\tau}_{k_i}(t).
  \label{eq:rigid}
\end{equation}\endgroup
A point now either fits its motion or does not, and the assignments are a segmentation of the scene with no further step.
The background needs no mask: it is the camera's motion, one more rigid motion like the rest. Once trajectories are
given, the left side of \cref{eq:rigid} is observed, and only the assignments and the motions remain unknown. Stacking the trajectories of
motion $k$ gives $\mathbf{X}_k = \mathbf{G}_k \mathbf{S}_k$, with its motion at every frame in $\mathbf{G}_k$ and its
canonical points, in homogeneous coordinates, in $\mathbf{S}_k$. Each block thus has rank at most 4, the structure of
multibody factorization~\cite{costeira1998multibody}, so the factorization can be computed rather than learned. \Cref{sec:em} computes it with EM, \cref{sec:ops} revises it, and
\cref{sec:scheme} fits Gaussians to the video (\cref{fig:overview}).

\subsection{Computing the factorization with EM}
\label{sec:em}
\begin{figure*}[t]
  \centering
  \includegraphics[width=\textwidth]{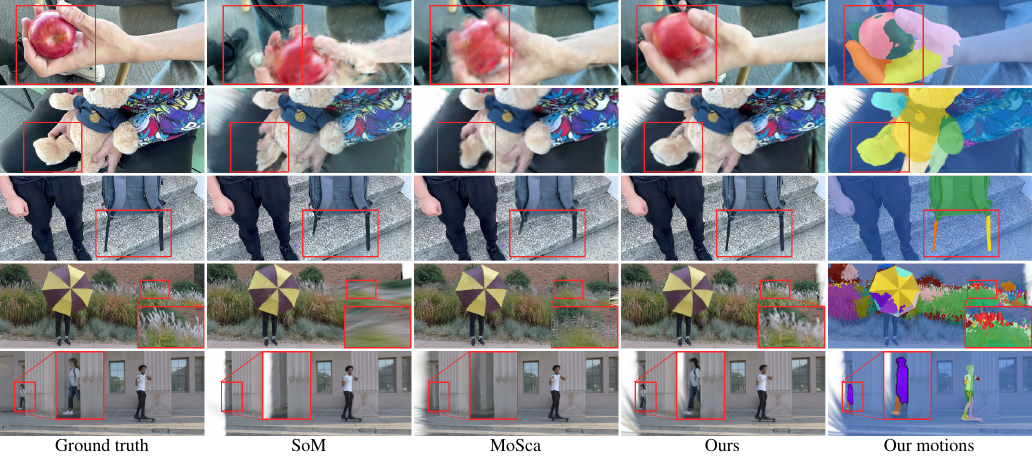}
  \vspace{-2mm}%
  \caption{Views of five scenes, held out except for backpack, whose single camera leaves only training views. Columns: ground truth, SoM, MoSca, \method, and the motions of \method, one colour per motion
  drawn over its rendering.}
  \label{fig:qual}
\vspace{-2mm}%
\end{figure*}
We take as canonical the frame in which most points are visible: $\mathbf{x}_i^c$ is the position
of point $i$ in that frame, where every motion is the identity. The assignments and motions should
predict each trajectory from its canonical point, so we minimise
\begingroup\setlength{\abovedisplayskip}{4pt plus 1pt minus 1pt}\setlength{\belowdisplayskip}{4pt plus 1pt minus 1pt}%
\begin{equation}
  E = \sum_i \sum_t w_i(t)\,\bigl\|\mathbf{x}_i(t) - \mathbf{R}_{k_i}(t)\,\mathbf{x}_i^c
      - \boldsymbol{\tau}_{k_i}(t)\bigr\|^2,
  \label{eq:energy}
\end{equation}\endgroup
where $w_i(t)$ is the tracker's confidence in point $i$ at frame $t$. Each unknown depends on the
other, but fixing one makes the other easy to solve, which is the setting of
EM~\cite{dempster1977em}. We use its hard form~\cite{celeux1992cem}: under Gaussian noise,
\cref{eq:energy} is the negative log-likelihood of the trajectories and their assignments, and each
point commits to its most likely motion.

We initialise the assignments with k-means on canonical position and velocity direction, and then
alternate two steps. With the assignments fixed, \cref{eq:energy} separates into one weighted
Procrustes problem per motion and frame, each with a closed-form solution: this is the M-step,
\begingroup\setlength{\abovedisplayskip}{4pt plus 1pt minus 1pt}\setlength{\belowdisplayskip}{4pt plus 1pt minus 1pt}%
\begin{equation}
  g_k(t) \leftarrow \argmin_{\mathbf{R},\boldsymbol{\tau}} \sum_{i:\,k_i=k} w_i(t)\,
  \bigl\|\mathbf{x}_i(t) - \mathbf{R}\,\mathbf{x}_i^c - \boldsymbol{\tau}\bigr\|^2.
  \label{eq:mstep}
\end{equation}\endgroup
With the motions fixed, it separates over points, and each point moves to the motion with the lowest
error over all frames: this is the E-step,
\begingroup\setlength{\abovedisplayskip}{4pt plus 1pt minus 1pt}\setlength{\belowdisplayskip}{4pt plus 1pt minus 1pt}%
\begin{equation}
  k_i \leftarrow \argmin_{k} \sum_t w_i(t)\,
  \bigl\|\mathbf{x}_i(t) - \mathbf{R}_k(t)\,\mathbf{x}_i^c - \boldsymbol{\tau}_k(t)\bigr\|^2.
  \label{eq:estep}
\end{equation}\endgroup
A point hidden in the canonical frame has no observed $\mathbf{x}_i^c$, so the E-step chooses it
together with the motion. Under motion $k$, the value that minimises \cref{eq:energy} is the
weighted mean of the trajectory with the motion undone, over its visible frames $\mathcal{V}_i$,
\begingroup\setlength{\abovedisplayskip}{4pt plus 1pt minus 1pt}\setlength{\belowdisplayskip}{4pt plus 1pt minus 1pt}%
\begin{equation}
  \mathbf{x}_i^c = \frac{\sum_{t \in \mathcal{V}_i} w_i(t)\,\mathbf{R}_k(t)^\top
    \bigl(\mathbf{x}_i(t) - \boldsymbol{\tau}_k(t)\bigr)}{\sum_{t \in \mathcal{V}_i} w_i(t)}.
  \label{eq:freepoint}
\end{equation}\endgroup
In terms of the blocks $\mathbf{X}_k$, the M-step fits $\mathbf{G}_k$ with $\mathbf{S}_k$ fixed, and
the E-step moves trajectories between blocks. Neither step can raise $E$, so EM descends to a local minimum.

We stop when the assignments no longer change. The output is an assignment $k_i$ per point and a motion $g_k(t)$
per $k$ and frame.
One motion has a known meaning: a static point moves only with the camera, so we call \emph{background} the
motion whose points the given cameras explain best.

\subsection{Revising the factorization}
\label{sec:ops}
EM holds two things fixed: the set of points and the number of motions. Requery changes the first, and
merge and split change the second. \Cref{fig:tools} shows each on real scenes.

\paragraph{Requery.}
\Cref{eq:energy} counts only the points that have a trajectory. A part the tracker missed adds nothing to the
energy, however wrong the model is there. A rendering of the current model shows where this happens. With $I_t$ a frame
and $\hat{I}_t$ its rendering, we keep the pixels of high error,
\begingroup\setlength{\abovedisplayskip}{4pt plus 1pt minus 1pt}\setlength{\belowdisplayskip}{4pt plus 1pt minus 1pt}%
\begin{equation}
  \begin{aligned}
  \epsilon_t(\mathbf{p}) &= \tfrac{1}{3}\bigl\|\hat{I}_t(\mathbf{p}) - I_t(\mathbf{p})\bigr\|_1,\\
  \Omega_t &= \bigl\{\mathbf{p} : \epsilon_t(\mathbf{p}) > \theta_t\bigr\},
  \end{aligned}
  \label{eq:errormap}
\end{equation}\endgroup
where $\theta_t$ is a high quantile of $\epsilon_t$ with a floor. Each frame receives queries in $\Omega_t$ in proportion
to its error, and the tracker follows each query through the video. We keep a new trajectory only if the rendering
error at its pixel is high on a fixed share of the frames where it is visible. A missing part is wrong wherever it is
seen, a rendering glitch only on a frame or two. A trajectory that follows an existing one is dropped. Each new point is
then offered to the motion that fits it best, the background included. It joins only if its error is no larger than
is typical for that motion's own points, over the whole video and in the frames around its query. The window matters
because a thin part can move differently for a few frames yet look right on average. Old points next to the
queries take the same windowed test. New points that no motion accepts and old points that fail this test are
grouped by \cref{eq:mstep,eq:estep} on their own, and their motions join the set, so $K$ grows. Requery never changes a
trajectory: it adds points and moves old ones between motions.

\begin{table*}[t]
  \caption{Novel view synthesis on SoM's five iPhone and seven NVIDIA scenes, with masked metrics, and tracking on all
  twelve iPhone scenes (MoSca: eleven). Masks: whether a method needs masks of what moves. Lower is better for EPE and LPIPS, higher otherwise. Best in bold. $^\dagger$Re-run by us
  without test-pose optimisation, with UniDepth instead of LiDAR on iPhone and its default Metric3D on NVIDIA.}
  \label{tab:main}
  \vspace{-1.5mm}%
  \centering
  \small
  \begin{tabular}{@{}ll ccc@{\hspace{3pt}\hspace{2\tabcolsep}}ccc@{\hspace{9pt}\hspace{2\tabcolsep}}ccc@{}}
    \toprule
    & & \multicolumn{6}{c}{iPhone} & \multicolumn{3}{c}{NVIDIA} \\
    \cmidrule(lr){3-8} \cmidrule(l){9-11}
    Method & Masks & EPE & AJ & PCK & PSNR & SSIM & LPIPS
      & PSNR & SSIM & LPIPS \\
    \midrule
    Marbles~\cite{stearns2024marbles} & \cmark & -- & -- & -- & -- & -- & -- & 23.46 & 0.77 & 0.093 \\
    MBGS~\cite{zhang2025motionblender} & \cmark & -- & -- & -- & 16.79 & 0.65 & 0.37 & -- & -- & -- \\
    SoM~\cite{wang2024shapeofmotion} & \cmark & 0.082 & 34.4 & 47.0 & 16.72 & 0.63 & 0.45 & 23.37 & 0.75 & 0.102 \\
    MoSca~\cite{lei2024mosca}$^\dagger$ & \cmark & 0.073 & 38.1 & 51.2 & 16.14 & 0.57 & 0.46 & 23.04 & 0.75 & 0.120 \\
    \method & \xmark & \textbf{0.058} & \textbf{41.0} & \textbf{53.2} & \textbf{18.05} & \textbf{0.66} & \textbf{0.34}
      & \textbf{23.83} & \textbf{0.80} & \textbf{0.089} \\
    \bottomrule
  \end{tabular}
\end{table*}
\paragraph{Merge.}
$K$ is set high so that every rigid part receives a motion, and most scenes end with more motions than rigid parts. The
background is the clearest case: EM often splits its points across several motions. The given cameras decide which
motions belong to it. A motion joins the background if the cameras alone explain its points within the error of the
background's own points. No other motion ever joins it. Two of the remaining motions should be merged when one rigid
motion explains the points of both. We fit that motion to the union of their points with \cref{eq:mstep} and measure
how much the error of each motion's points rises over that motion's own fit. The scale of this rise comes from the
background itself. We cut two nearby patches from its points, with as many points as the two motions under test, and
fit them together. They share one motion, so their rise is only noise. A motion objects to a merge if its rise exceeds
that of every such pair and the background's own error. The pair with the smallest rise merges first, and we repeat
until no pair passes. When two groups are tested, any motion they contain can object, not only the last one added.
Small differences therefore cannot add up along a chain.

\vspace{-1mm}%
\paragraph{Split.}
EM can also give one motion to two parts that move differently. This happens when they were close at initialisation,
or when one of them is too small to win a motion of its own. A motion is split when two rigid motions explain its points
better than one, by more than noise alone would allow. We run \cref{eq:mstep,eq:estep} with two motions on its points,
several times from different first guesses, such as its worst-fitting points against the rest. Each run gives a cut. Its gain is the drop in the
error of the new side once that side receives its own motion. Noise sets the scale again, this time from the motion
itself. We build copies of its points that hold a single motion by construction: each point sits at the single motion's
prediction plus its own residual under the cut, with a random sign. The same procedure runs on every copy. A cut passes
only if its gain exceeds that of every copy with a new side of similar size, and the passing cut with the largest gain
is made. The new motions are tested in turn, so $K$ grows without a cap.

\subsection{Training the scene}
\label{sec:scheme}
The factorization becomes a renderable scene with one 3D Gaussian~\cite{kerbl20233dgs} per point. Its centre
$\boldsymbol{\mu}_i$ starts at the point's canonical position and moves with the point's motion: at frame $t$ it sits at
$\mathbf{R}_{k_i}(t)\boldsymbol{\mu}_i + \boldsymbol{\tau}_{k_i}(t)$, and its orientation turns with
$\mathbf{R}_{k_i}(t)$. The assignment is one-hot and gradients never change it, so a Gaussian never blends
motions. Training refines every motion frame by frame, except the background's, which stays the given camera
motion.

We train with the losses of Shape of Motion~\cite{wang2024shapeofmotion}: a photometric loss, a loss that keeps each Gaussian on its point's 2D track, a depth loss and
a prior that keeps the motions smooth in time. Density control clones, splits and prunes Gaussians, and a new Gaussian
copies its parent's motion. The number of motions and the assignment of every point therefore change only
through requery, merge and split.

The tools run in a fixed order. Split and merge first revise the factorization that EM returns, before any rendering
exists. A short first stage then gives the rendering that requery needs. Merge runs once more, and every motion is refit
to its points with robust weights, alternating \cref{eq:mstep,eq:freepoint}. The final stage then trains the scene from
this factorization, starting again from one Gaussian per point.
\section{Experiments}
\label{sec:experiments}
\begin{table*}[t]
  \caption{SE(3) motion segmentation on MoSE3's 55 iTACO clips and the full public MOVi-F validation split (147 scenes). Lower is
  better for RRE and ADD, higher for the rest. Best in bold, second underlined. All rows are our runs. $^\ast$Concurrent work.}
  \label{tab:seg}
  \centering
  \small
  \setlength{\tabcolsep}{3pt}
  \resizebox{\textwidth}{!}{%
  \begin{tabular}{@{}l *{4}{c}@{\hspace{3pt}\hspace{2\tabcolsep}}*{4}{c}@{\hspace{9pt}\hspace{2\tabcolsep}}*{4}{c}@{\hspace{3pt}\hspace{2\tabcolsep}}*{4}{c}@{\hspace{9pt}\hspace{2\tabcolsep}}cc@{}}
    \toprule
    & \multicolumn{8}{c}{Per-pixel SE(3)} & \multicolumn{8}{c}{Object / part-level SE(3)} & \multicolumn{2}{c}{Assignment} \\
    \cmidrule(lr){2-9} \cmidrule(lr){10-17} \cmidrule(l){18-19}
    & \multicolumn{4}{c}{iTACO} & \multicolumn{4}{c}{MOVi-F} & \multicolumn{4}{c}{iTACO} & \multicolumn{4}{c}{MOVi-F} & \multicolumn{2}{c}{MOVi-F} \\
    \cmidrule(lr){2-5} \cmidrule(lr){6-9} \cmidrule(lr){10-13} \cmidrule(lr){14-17} \cmidrule(l){18-19}
    Method & RRE & ADD & AUC$_{r}$ & AUC$_{\mathrm{ADD}}$ & RRE & ADD & AUC$_{r}$ & AUC$_{\mathrm{ADD}}$ & RRE & ADD & AUC$_{\mathrm{ADD}}$ & IoU & RRE & ADD & AUC$_{\mathrm{ADD}}$ & IoU & ARI & mIoU \\
    \midrule
    Segment Any Motion~\cite{huang2025segmentanymotion} & -- & -- & -- & -- & -- & -- & -- & -- & -- & -- & -- & 0.477 & -- & -- & -- & -- & 0.579 & 0.500 \\
    Trace Anything~\cite{liu2026traceanything} & 44.72 & 35.48 & 0.347 & 0.252 & 45.66 & 148.45 & 0.345 & 0.127 & 37.53 & 31.95 & 0.353 & 0.143 & 33.78 & 109.90 & \textbf{0.288} & 0.262 & 0.299 & 0.309 \\
    4RC~\cite{luo20264rc} & 28.54 & 20.88 & 0.473 & 0.416 & 34.89 & \underline{71.06} & 0.455 & \underline{0.185} & 20.22 & 16.93 & 0.539 & 0.337 & 25.36 & \textbf{60.84} & \underline{0.282} & 0.456 & 0.479 & 0.467 \\
    Track4World~\cite{lu2026track4world} & 28.18 & 24.67 & 0.446 & 0.314 & 37.31 & 99.20 & 0.392 & 0.091 & 19.30 & 20.21 & 0.395 & 0.357 & \underline{20.42} & \underline{69.22} & 0.190 & 0.292 & 0.430 & 0.340 \\
    MoSE3$^{\ast}$~\cite{cheng2026mose3} & \textbf{12.30} & \textbf{11.37} & \textbf{0.741} & \textbf{0.594} & \textbf{19.46} & \textbf{62.98} & \textbf{0.600} & 0.131 & \textbf{10.76} & \textbf{10.49} & \underline{0.625} & \textbf{0.784} & \textbf{19.38} & 73.39 & 0.161 & \textbf{0.897} & \underline{0.743} & \textbf{0.813} \\
    \midrule
    \method, MoSE3's tracks & \underline{16.81} & \underline{14.15} & \underline{0.707} & 0.566 & \underline{22.76} & 92.69 & \textbf{0.600} & 0.142 & \underline{13.31} & \underline{12.15} & 0.608 & \underline{0.701} & 21.14 & 84.61 & 0.170 & \underline{0.616} & 0.729 & 0.623 \\
    \method & 26.72 & 24.88 & 0.634 & \underline{0.571} & 28.31 & 102.51 & \underline{0.562} & \textbf{0.195} & 23.58 & 20.27 & \textbf{0.630} & 0.661 & 27.43 & 104.95 & 0.223 & 0.596 & \textbf{0.775} & \underline{0.627} \\
    \bottomrule
  \end{tabular}}
\end{table*}

For views and tracks we follow Shape of Motion (SoM)~\cite{wang2024shapeofmotion}: the iPhone scenes of
DyCheck~\cite{gao2022dycheck} at full resolution, and the seven NVIDIA scenes~\cite{yoon2020novel}, trained on one camera
and tested on the rest. We report PSNR, SSIM and LPIPS inside the covisibility masks, and the 3D error (EPE) and 2D
accuracy (AJ, PCK) of the iPhone keypoints. Motions are scored on MOVi-F~\cite{greff2022kubric} and
iTACO~\cite{peng2026itaco}.

Point tracks come from AllTracker~\cite{harley2025alltracker} and depth from ARROW~\cite{fradlin2026arrow}, aligned to the
sparse points of the given cameras by one scale and shift per video. The cameras are SoM's COLMAP reconstruction on
iPhone and a static camera on NVIDIA. EM compares trajectories in 3D. The tools compare them after projection, in pixels,
where depth noise matters less. We start with $K=20$ motions, run the first stage for 1,500 steps and the final one for
500 epochs, and give every threshold in the supplement.

\subsection{View synthesis and tracking}
\label{sec:comparison}

\Cref{tab:main} compares \method with four methods. SoM, MoSca and MBGS blend several motions in every Gaussian, and
Marbles moves each Gaussian on its own. \method is best on every metric of \cref{tab:main}, by a wide margin on iPhone and in tracking. On iPhone it reaches 18.05 dB against 16.72 for
SoM and 16.14 for MoSca, and on NVIDIA 23.83 dB against 23.37 and 23.04. The gain does not come from a freer motion
model. Each Gaussian follows one rigid motion, while the other methods blend
several motions or move each Gaussian freely. The same model gives tracks for free, since a point moves with its motion. On the annotated
iPhone keypoints, \method has a 3D error of 0.058\,m and a 2D accuracy of 41.0 AJ and 53.2 PCK. SoM, trained with the
same losses, reaches 0.082\,m, 34.4 AJ and 47.0 PCK.

The difference is just as clear in \cref{fig:qual}, which shows five scenes. Without masks, every motion in the scene competes for its points, not
only the objects a segmenter was asked to find. Small and distant movers, such as the woman walking behind a pillar in
Skating, therefore get a motion too. A method whose foreground mask leaves them out treats them as static. Swaying grass,
as behind the umbrella in Umbrella, is just as hard to outline with a mask, and \method gives it motions of its own.
Split, in turn, separates distinct rigid parts such as fingers or the teddy's legs, which keeps them sharp where the
baselines blur them. Motion shows far better in video than in still frames, and we strongly encourage the
reader to watch the supplementary video.

\Cref{tab:main} keeps to SoM's protocol, with the given cameras, monocular depth and no test-pose optimisation. SoM, MBGS
and Marbles report under it, MBGS in its own paper and Marbles in SoM's, and we take their numbers. We
rerun MoSca with monocular depth instead of LiDAR and without its test-pose optimisation, like every other row. With both,
its paper reports 18.40\,dB, and its own ablation loses 1.7 to 2.3\,dB when monocular depth replaces LiDAR. Other recent methods stay out of \cref{tab:main} for one of three reasons. ViDAR~\cite{nazarczuk2025vidar},
CAT4D~\cite{wu2025cat4d} and CogNVS~\cite{chen2025cognvs} add a generative prior, which fills unseen views with a model
trained on other data. MotionScale~\cite{zhou2026motionscale} does not state its scenes, resolution or masks, and its
released code has no iPhone or NVIDIA setup. Ground4D~\cite{zhao2026ground4d} and Motion4D~\cite{zhou2025motion4d} have
released no code. Motion4D claims SoM's cameras and depth but does not state its masks, scenes or resolution.
Ground4D does not state its cameras or test poses. The supplement defines our
protocol in full and lists the other methods' numbers with the settings each states. We encourage future work to do the same for fair comparisons, as these settings move PSNR as
much as the gaps between methods.

\subsection{SE(3) motion segmentation}
\label{sec:segmentation}

A rendering can look right with the wrong motions, so we also score the motions themselves (\cref{tab:seg}).
MOVi-F~\cite{greff2022kubric} renders rigid objects moving in a scene, and iTACO~\cite{peng2026itaco} shows articulated
objects, such as a microwave whose door turns about its hinge. \method runs on both exactly as for reconstruction, with the
same pipeline and settings and the datasets' cameras, and we score the factorization that its final reconstruction starts from.
Each object or part is matched by IoU to one motion, and that motion is scored by its rotation error (RRE, in degrees) and
translation error (ADD, in cm) against the true motion, per pixel and per object. On MOVi-F the object behind every track is known, so we also compare the track assignments with the
objects by the adjusted Rand index (ARI), which counts the background too, and the mean IoU. We compare with Segment Any Motion~\cite{huang2025segmentanymotion},
which segments long tracks with a learned model, with Trace Anything~\cite{liu2026traceanything}, 4RC~\cite{luo20264rc}
and Track4World~\cite{lu2026track4world}, feed-forward networks that predict the 3D trajectory of every pixel, and with
the concurrent MoSE3~\cite{cheng2026mose3}, which predicts rigid groups directly.

\begin{figure*}[t]
  \centering
  \includegraphics[width=1.0\textwidth,trim=0 19.4 0 0,clip]{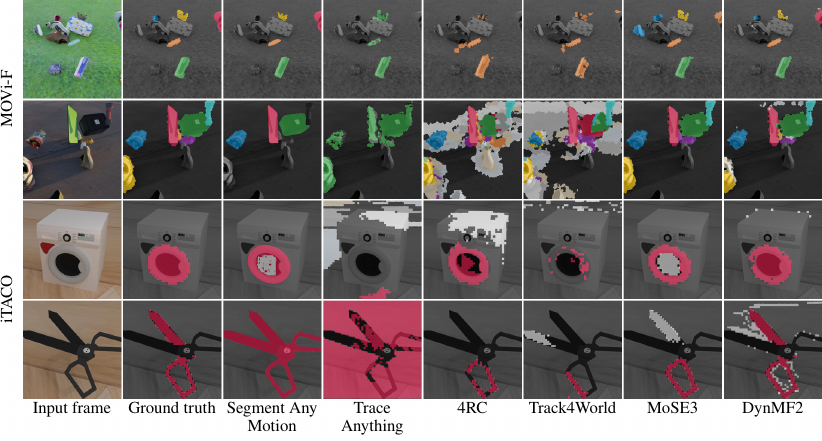}\\[0.3pt]
  \setlength{\unitlength}{\textwidth}\begin{picture}(1,0.017)\put(0.0876,0){\makebox(0,0)[b]{\scriptsize\strut Input frame}}\put(0.2093,0){\makebox(0,0)[b]{\scriptsize\strut Ground truth}}\put(0.3311,0){\makebox(0,0)[b]{\scriptsize\strut Segment Any Motion}}\put(0.4529,0){\makebox(0,0)[b]{\scriptsize\strut Trace Anything}}\put(0.5748,0){\makebox(0,0)[b]{\scriptsize\strut 4RC}}\put(0.6965,0){\makebox(0,0)[b]{\scriptsize\strut Track4World}}\put(0.8184,0){\makebox(0,0)[b]{\scriptsize\strut MoSE3}}\put(0.9402,0){\makebox(0,0)[b]{\scriptsize\strut \method}}\end{picture}
  \vspace{-3mm}%
  \caption{Motion segmentation on two MOVi-F scenes (top) and two iTACO clips (bottom): input frame, ground truth (the moving
  objects in MOVi-F, the moving part in iTACO), Segment Any Motion, Trace Anything, 4RC, Track4World, MoSE3 and the motions of
  \method, coloured to match the ground truth, with the static scene left uncoloured and unmatched groups in light grey. MoSE3
  is concurrent work.}
  \label{fig:segqual}
\vspace{-0.5mm}%
\end{figure*}

On MOVi-F, \method has the best ARI, 0.775, against 0.743 for MoSE3 and 0.579 for Segment Any Motion, and the second mIoU,
0.627, behind MoSE3's 0.813. On iTACO its part IoU of 0.661 trails MoSE3's 0.784 and leads every other method. MoSE3 also
has the lowest rotation and translation errors almost everywhere, as it is built for this task. On iTACO, much of the gap is the tracker.
Given MoSE3's 3D tracks and estimated cameras but not its learned rigidity embeddings, which do its grouping, \method
reaches 0.701 IoU there, and
its per-pixel errors fall by more than a third. \Cref{fig:segqual} shows examples.

\begin{figure}[t]
\centering
\includegraphics[width=\linewidth]{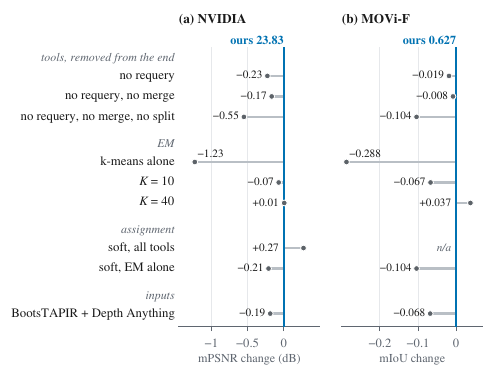}
\vspace{-2.5mm}%
\caption{\textbf{Ablations.} Each row changes one part of \method and shows the change in score. (a) NVIDIA, test mPSNR over seven scenes. (b) MOVi-F, mIoU of the motions, 147 validation scenes. Tools are removed from the last one back. One run per scene.}
\label{fig:ablations}
\end{figure}

\subsection{Ablations}
\label{sec:ablations}

\Cref{fig:ablations} changes one part of \method at a time, scored by test PSNR after full training on NVIDIA and by the
mIoU of its motions on MOVi-F. Most of the gain comes from the pipeline, not its inputs: on SoM's inputs, BootsTAPIR tracks (MoSca's
too) and Depth Anything depth, \method loses only 0.19\,dB on NVIDIA, where it still beats every other method, and 0.07 mIoU
on MOVi-F.

The EM matters most. Grouping the tracks once by k-means, with no EM and no tools, and fitting each group, the common start of current methods,
costs 1.23\,dB and 0.288 mIoU, and the EM alone recovers 0.68\,dB and 0.184 mIoU of this. Each tool then adds to the EM. Split recovers most of what the EM alone loses, 0.38 of
0.55\,dB and 0.096 of 0.104 mIoU. Merge joins the pieces of the background at almost no cost in either score, and
requery adds 0.23\,dB. Only the full pipeline recovers small details and distant movers.

Split and merge pull the number of motions toward the scene without erasing the start. On NVIDIA, starting from 10 or 40
motions instead of 20 changes PSNR by less than 0.1\,dB. On MOVi-F, scenes end with 14, 16 and 20 motions on average from
starts of 10, 20 and 40, against 7.5 true ones. More motions miss fewer objects but cut more of them, so 40 gains
0.037 mIoU and 10 loses 0.067.

Hard assignment is competitive with a blend. Blending the EM's motions, as the bases of Shape of Motion do, stays
0.21\,dB below \method, and blending the motions the tools revised gains 0.27\,dB over holding them hard. That gain sits
where the scene bends: Jumping's bodies (+1.14\,dB), the grass in Umbrella (+0.45) and the skater's legs (+0.32), while
the rigid scenes do not change substantially. The advantage of soft blending in these scenes does not count against strict rigid
assignment. Rather, it marks a limitation of the framework and points to promising future work: $\mathrm{SE}(3)$ motions
could be hierarchical, with soft blending allowed only between a parent motion and its children, and only at their
border.
\vspace{1.7mm}%
\section{Conclusion}
\label{sec:conclusion}

\method computes the motion factorization of a dynamic scene instead of learning it through the renderer. EM on 3D
trajectories gives each point a single rigid motion, and requery, merge and split revise the set of motions and the points
they hold. The scene needs no masks, each Gaussian moves rigidly with one motion, and the same motions segment what moves.
\method leads every metric of view synthesis and tracking in \cref{tab:main}, and its motions, never trained to segment, are competitive
with methods trained to do so.

Depth is now the limit. With sensor depth the same pipeline renders better, and with the exact 3D trajectories of MOVi-F,
EM recovers the rigid parts almost exactly (supplement). The tools measure their errors in pixels only to shield the
factorization from depth errors. The factorization can also give something back: a scene moves by a few $\mathrm{SE}(3)$
transformations, a structure that point tracks lack. MoSE3~\cite{cheng2026mose3} learns an $\mathrm{SE}(3)$ motion and a
track for every pixel. A tracker could instead learn only the few motions of a scene and the assignment of each point, with
this EM inside its training loop. Trajectories and motions would then sharpen each other: better trajectories give cleaner
motions, and cleaner motions show the tracker where it went wrong.
{}
\par
{
    \small
    \renewcommand*{\backref}[1]{\unskip\unskip~\mbox{#1}}%
    \microtypesetup{protrusion=true,expansion=true}%
    \bibliographystyle{ieeenat_fullname}
    \bibliography{main}

\begin{thebibliography}{111}
\providecommand{\natexlab}[1]{#1}
\providecommand{\url}[1]{\texttt{#1}}
\expandafter\ifx\csname urlstyle\endcsname\relax
  \providecommand{\doi}[1]{doi: #1}\else
  \providecommand{\doi}{doi: \begingroup \urlstyle{rm}\Url}\fi

\bibitem[Akhter et~al.(2011)Akhter, Sheikh, Khan, and Kanade]{akhter2011trajectory}
Ijaz Akhter, Yaser Sheikh, Sohaib Khan, and Takeo Kanade.
\newblock Trajectory space: A dual representation for nonrigid structure from motion.
\newblock \emph{IEEE TPAMI}, 33\penalty0 (7):\penalty0 1442--1456, 2011.

\bibitem[Allshire et~al.(2025)Allshire, Choi, Zhang, McAllister, Zhang, Kim, Darrell, Abbeel, Malik, and Kanazawa]{allshire2025videomimic}
Arthur Allshire, Hongsuk Choi, Junyi Zhang, David McAllister, Anthony Zhang, Chung~Min Kim, Trevor Darrell, Pieter Abbeel, Jitendra Malik, and Angjoo Kanazawa.
\newblock Visual imitation enables contextual humanoid control.
\newblock In \emph{Conference on Robot Learning (CoRL)}, pages 794--815, 2025.

\bibitem[Ayer and Sawhney(1995)]{ayer1995layered}
Serge Ayer and Harpreet~S. Sawhney.
\newblock Layered representation of motion video using robust maximum-likelihood estimation of mixture models and {MDL} encoding.
\newblock In \emph{ICCV}, pages 777--784, 1995.

\bibitem[Bregler et~al.(2000)Bregler, Hertzmann, and Biermann]{bregler2000recovering}
Christoph Bregler, Aaron Hertzmann, and Henning Biermann.
\newblock Recovering non-rigid {3D} shape from image streams.
\newblock In \emph{CVPR}, pages 690--696, 2000.

\bibitem[Brox and Malik(2010)]{brox2010object}
Thomas Brox and Jitendra Malik.
\newblock Object segmentation by long term analysis of point trajectories.
\newblock In \emph{ECCV}, pages 282--295, 2010.

\bibitem[Broxton et~al.(2020)Broxton, Flynn, Overbeck, Erickson, Hedman, DuVall, Dourgarian, Busch, Whalen, and Debevec]{broxton2020immersive}
Michael Broxton, John Flynn, Ryan Overbeck, Daniel Erickson, Peter Hedman, Matthew DuVall, Jason Dourgarian, Jay Busch, Matt Whalen, and Paul Debevec.
\newblock Immersive light field video with a layered mesh representation.
\newblock \emph{ACM TOG}, 39\penalty0 (4):\penalty0 86:1--86:15, 2020.

\bibitem[Celeux and Govaert(1992)]{celeux1992cem}
Gilles Celeux and G{\'e}rard Govaert.
\newblock A classification {EM} algorithm for clustering and two stochastic versions.
\newblock \emph{Computational Statistics \& Data Analysis}, 14\penalty0 (3):\penalty0 315--332, 1992.

\bibitem[Chen et~al.(2025)Chen, Khurana, and Ramanan]{chen2025cognvs}
Kaihua Chen, Tarasha Khurana, and Deva Ramanan.
\newblock Reconstruct, inpaint, test-time finetune: Dynamic novel-view synthesis from monocular videos.
\newblock In \emph{NeurIPS}, pages 53255--53285, 2025.
\newblock arXiv:2507.12646.

\bibitem[Cheng et~al.(2026)Cheng, Li, Xia, Cai, Du, and Wang]{cheng2026mose3}
Jiahuan Cheng, Zhiyi Li, Tian Xia, Ruojin Cai, Yilun Du, and Qianqian Wang.
\newblock {MoSE3}: Learning world-space {SE(3)} at every pixel.
\newblock In \emph{NeurIPS}, 2026.

\bibitem[Costeira and Kanade(1998)]{costeira1998multibody}
Jo{\~a}o~Paulo Costeira and Takeo Kanade.
\newblock A multibody factorization method for independently moving objects.
\newblock \emph{IJCV}, 29\penalty0 (3):\penalty0 159--179, 1998.

\bibitem[Delong et~al.(2012)Delong, Osokin, Isack, and Boykov]{delong2012labelcosts}
Andrew Delong, Anton Osokin, Hossam~N. Isack, and Yuri Boykov.
\newblock Fast approximate energy minimization with label costs.
\newblock \emph{IJCV}, 96\penalty0 (1):\penalty0 1--27, 2012.

\bibitem[Dempster et~al.(1977)Dempster, Laird, and Rubin]{dempster1977em}
Arthur~P. Dempster, Nan~M. Laird, and Donald~B. Rubin.
\newblock Maximum likelihood from incomplete data via the {EM} algorithm.
\newblock \emph{Journal of the Royal Statistical Society: Series B}, 39\penalty0 (1):\penalty0 1--22, 1977.

\bibitem[Doersch et~al.(2024)Doersch, Luc, Yang, Gokay, Koppula, Gupta, Heyward, Rocco, Goroshin, Carreira, and Zisserman]{doersch2024bootstap}
Carl Doersch, Pauline Luc, Yi Yang, Dilara Gokay, Skanda Koppula, Ankush Gupta, Joseph Heyward, Ignacio Rocco, Ross Goroshin, Jo{\~a}o Carreira, and Andrew Zisserman.
\newblock {BootsTAP}: Bootstrapped training for tracking-any-point.
\newblock In \emph{ACCV}, pages 483--500, 2024.

\bibitem[Elhamifar and Vidal(2013)]{elhamifar2013sparse}
Ehsan Elhamifar and Ren{\'e} Vidal.
\newblock Sparse subspace clustering: Algorithm, theory, and applications.
\newblock \emph{IEEE TPAMI}, 35\penalty0 (11):\penalty0 2765--2781, 2013.

\bibitem[Feng et~al.(2025)Feng, Zhang, Wang, Ye, Yu, Black, Darrell, and Kanazawa]{feng2025st4rtrack}
Haiwen Feng, Junyi Zhang, Qianqian Wang, Yufei Ye, Pengcheng Yu, Michael~J. Black, Trevor Darrell, and Angjoo Kanazawa.
\newblock {St4RTrack}: Simultaneous {4D} reconstruction and tracking in the world.
\newblock In \emph{ICCV}, pages 8503--8513, 2025.

\bibitem[Fradlin et~al.(2026)Fradlin, Schmidt, Piekenbrinck, Knaebel, Martin~Garcia, and Leibe]{fradlin2026arrow}
Ilya Fradlin, Christian Schmidt, Jens Piekenbrinck, Karim Knaebel, Gonzalo Martin~Garcia, and Bastian Leibe.
\newblock {ARROW}: Arbitrary reconstruction and tracking of {4D} observations in the wild.
\newblock \emph{arXiv preprint arXiv:2610.01314}, 2026.

\bibitem[Gao et~al.(2021)Gao, Saraf, Kopf, and Huang]{gao2021dynnerf}
Chen Gao, Ayush Saraf, Johannes Kopf, and Jia-Bin Huang.
\newblock Dynamic view synthesis from dynamic monocular video.
\newblock In \emph{ICCV}, pages 5692--5701, 2021.

\bibitem[Gao et~al.(2022)Gao, Li, Tulsiani, Russell, and Kanazawa]{gao2022dycheck}
Hang Gao, Ruilong Li, Shubham Tulsiani, Bryan Russell, and Angjoo Kanazawa.
\newblock Monocular dynamic view synthesis: A reality check.
\newblock In \emph{NeurIPS}, pages 33768--33780, 2022.

\bibitem[Greff et~al.(2022)Greff, Belletti, Beyer, Doersch, Du, Duckworth, Fleet, Gnanapragasam, Golemo, Herrmann, Kipf, Kundu, Lagun, Laradji, Liu, Meyer, Miao, Nowrouzezahrai, Oztireli, Pot, Radwan, Rebain, Sabour, Sajjadi, Sela, Sitzmann, Stone, Sun, Vora, Wang, Wu, Yi, Zhong, and Tagliasacchi]{greff2022kubric}
Klaus Greff, Francois Belletti, Lucas Beyer, Carl Doersch, Yilun Du, Daniel Duckworth, David~J. Fleet, Dan Gnanapragasam, Florian Golemo, Charles Herrmann, Thomas Kipf, Abhijit Kundu, Dmitry Lagun, Issam Laradji, Hsueh-Ti~(Derek) Liu, Henning Meyer, Yishu Miao, Derek Nowrouzezahrai, Cengiz Oztireli, Etienne Pot, Noha Radwan, Daniel Rebain, Sara Sabour, Mehdi S.~M. Sajjadi, Matan Sela, Vincent Sitzmann, Austin Stone, Deqing Sun, Suhani Vora, Ziyu Wang, Tianhao Wu, Kwang~Moo Yi, Fangcheng Zhong, and Andrea Tagliasacchi.
\newblock Kubric: A scalable dataset generator.
\newblock In \emph{CVPR}, pages 3749--3761, 2022.

\bibitem[Guo et~al.(2026{\natexlab{a}})Guo, Hsu, Ding, and Zhang]{guo2026usplat4d}
Fengzhi Guo, Chih-Chuan Hsu, Sihao Ding, and Cheng Zhang.
\newblock Uncertainty matters in dynamic {G}aussian splatting for monocular {4D} reconstruction.
\newblock In \emph{ICLR}, 2026{\natexlab{a}}.
\newblock arXiv:2510.12768.

\bibitem[Guo et~al.(2026{\natexlab{b}})Guo, Zhao, Zhao, Fu, Zhuang, Huang, Li, and Zou]{guo2026articulat3d}
Lijun Guo, Haoyu Zhao, Xingyue Zhao, Rong Fu, Linghao Zhuang, Siteng Huang, Zhongyu Li, and Hua Zou.
\newblock {Articulat3D}: Reconstructing articulated digital twins from monocular videos with geometric and motion constraints.
\newblock In \emph{ECCV}, pages 58--76, 2026{\natexlab{b}}.
\newblock arXiv:2603.11606.

\bibitem[Han et~al.(2026)Han, Benarroch~Jedlicki, Lei, and Yang]{han2026xm4d}
Haoyu Han, Jack Benarroch~Jedlicki, Jiahui Lei, and Heng Yang.
\newblock {XM4D}: {4D} reconstruction through certifiable geometric optimization.
\newblock \emph{arXiv preprint arXiv:2611.07902}, 2026.

\bibitem[Harley et~al.(2025)Harley, You, Sun, Zheng, Raghuraman, Gu, Liang, Chu, Dave, You, Ambrus, Fragkiadaki, and Guibas]{harley2025alltracker}
Adam~W. Harley, Yang You, Xinglong Sun, Yang Zheng, Nikhil Raghuraman, Yunqi Gu, Sheldon Liang, Wen-Hsuan Chu, Achal Dave, Suya You, Rares Ambrus, Katerina Fragkiadaki, and Leonidas~J. Guibas.
\newblock {AllTracker}: Efficient dense point tracking at high resolution.
\newblock In \emph{ICCV}, pages 5253--5262, 2025.

\bibitem[Hu et~al.(2026)Hu, Cheng, Yu, Guo, and Wang]{hu2026vggt4d}
Yu Hu, Chong Cheng, Sicheng Yu, Xiaoyang Guo, and Hao Wang.
\newblock {VGGT4D}: Mining motion cues in visual geometry transformers for {4D} scene reconstruction.
\newblock In \emph{Proceedings of the IEEE/CVF Conference on Computer Vision and Pattern Recognition (CVPR) Findings}, pages 414--424, 2026.

\bibitem[Huang et~al.(2019)Huang, Yang, Zhao, Lai, and Hu]{huang2019clusterslam}
Jiahui Huang, Sheng Yang, Zishuo Zhao, Yu-Kun Lai, and Shi-Min Hu.
\newblock {ClusterSLAM}: A {SLAM} backend for simultaneous rigid body clustering and motion estimation.
\newblock In \emph{ICCV}, pages 5874--5883, 2019.

\bibitem[Huang et~al.(2020)Huang, Yang, Mu, and Hu]{huang2020clustervo}
Jiahui Huang, Sheng Yang, Tai-Jiang Mu, and Shi-Min Hu.
\newblock {ClusterVO}: Clustering moving instances and estimating visual odometry for self and surroundings.
\newblock In \emph{CVPR}, pages 2165--2174, 2020.

\bibitem[Huang et~al.(2025)Huang, Zheng, Xu, Keutzer, Zhang, Kanazawa, and Wang]{huang2025segmentanymotion}
Nan Huang, Wenzhao Zheng, Chenfeng Xu, Kurt Keutzer, Shanghang Zhang, Angjoo Kanazawa, and Qianqian Wang.
\newblock Segment any motion in videos.
\newblock In \emph{CVPR}, pages 3406--3416, 2025.
\newblock arXiv:2503.22268.

\bibitem[Huang et~al.(2024)Huang, Sun, Yang, Lyu, Cao, and Qi]{huang2024scgs}
Yi-Hua Huang, Yang-Tian Sun, Ziyi Yang, Xiaoyang Lyu, Yan-Pei Cao, and Xiaojuan Qi.
\newblock {SC-GS}: Sparse-controlled {G}aussian splatting for editable dynamic scenes.
\newblock In \emph{CVPR}, pages 4220--4230, 2024.

\bibitem[Isack and Boykov(2012)]{isack2012pearl}
Hossam Isack and Yuri Boykov.
\newblock Energy-based geometric multi-model fitting.
\newblock \emph{IJCV}, 97\penalty0 (2):\penalty0 123--147, 2012.

\bibitem[Kang et~al.(2025)Kang, Park, Lee, and Lee]{kang2025groupedfourdgs}
Taeho Kang, Jaeyeon Park, Kyungjin Lee, and Youngki Lee.
\newblock Clustered error correction with grouped {4D} {G}aussian splatting.
\newblock In \emph{SIGGRAPH Asia Conference Papers}, pages 1--12, 2025.
\newblock arXiv:2511.16112.

\bibitem[Karaev et~al.(2025)Karaev, Makarov, Wang, Neverova, Vedaldi, and Rupprecht]{karaev2025cotracker3}
Nikita Karaev, Yuri Makarov, Jianyuan Wang, Natalia Neverova, Andrea Vedaldi, and Christian Rupprecht.
\newblock {CoTracker3}: Simpler and better point tracking by pseudo-labelling real videos.
\newblock In \emph{ICCV}, pages 6013--6022, 2025.

\bibitem[Karazija et~al.(2024)Karazija, Laina, Rupprecht, and Vedaldi]{karazija2024trajectories}
Laurynas Karazija, Iro Laina, Christian Rupprecht, and Andrea Vedaldi.
\newblock Learning segmentation from point trajectories.
\newblock In \emph{NeurIPS}, pages 112573--112597, 2024.

\bibitem[Kerbl et~al.(2023)Kerbl, Kopanas, Leimk{\"u}hler, and Drettakis]{kerbl20233dgs}
Bernhard Kerbl, Georgios Kopanas, Thomas Leimk{\"u}hler, and George Drettakis.
\newblock {3D} {G}aussian splatting for real-time radiance field rendering.
\newblock \emph{ACM TOG}, 42\penalty0 (4):\penalty0 1--14, 2023.

\bibitem[Kim et~al.(2026)Kim, Kim, and Han]{kim2026gp4dgs}
Mijeong Kim, Jungtaek Kim, and Bohyung Han.
\newblock {GP-4DGS}: Probabilistic {4D} {G}aussian splatting from monocular video via variational {G}aussian processes.
\newblock In \emph{CVPR}, pages 33394--33403, 2026.
\newblock arXiv:2604.02915.

\bibitem[Kratimenos et~al.(2024)Kratimenos, Lei, and Daniilidis]{kratimenos2023dynmf}
Agelos Kratimenos, Jiahui Lei, and Kostas Daniilidis.
\newblock {DynMF}: Neural motion factorization for real-time dynamic view synthesis with {3D} {G}aussian splatting.
\newblock In \emph{ECCV}, pages 252--269. Springer, 2024.

\bibitem[Kumar et~al.(2017)Kumar, Dai, and Li]{kumar2017spatio}
Suryansh Kumar, Yuchao Dai, and Hongdong Li.
\newblock Spatio-temporal union of subspaces for multi-body non-rigid structure-from-motion.
\newblock \emph{PR}, 71:\penalty0 428--443, 2017.

\bibitem[Lee et~al.(2026)Lee, Choi, and Cho]{lee2026mogaf}
Junmyeong Lee, Hoseung Choi, and Minsu Cho.
\newblock Space-time forecasting of dynamic scenes with motion-aware {G}aussian grouping.
\newblock In \emph{CVPR}, pages 41033--41043, 2026.
\newblock arXiv:2602.21668.

\bibitem[Lei et~al.(2025)Lei, Weng, Harley, Guibas, and Daniilidis]{lei2024mosca}
Jiahui Lei, Yijia Weng, Adam~W. Harley, Leonidas Guibas, and Kostas Daniilidis.
\newblock {MoSca}: Dynamic {G}aussian fusion from casual videos via {4D} motion scaffolds.
\newblock In \emph{CVPR}, pages 6165--6177, 2025.
\newblock arXiv:2405.17421.

\bibitem[Li et~al.(2026)Li, Gu, Chen, Qiu, and Sun]{li2026gaussiansequences}
Can Li, Jie Gu, Jingmin Chen, Fangzhou Qiu, and Lei Sun.
\newblock {Gaussian Sequences with Multi-Scale Dynamics for 4D Reconstruction from Monocular Casual Videos}.
\newblock \emph{arXiv preprint arXiv:2602.13806}, 2026.

\bibitem[Li et~al.(2021)Li, Niklaus, Snavely, and Wang]{li2021nsff}
Zhengqi Li, Simon Niklaus, Noah Snavely, and Oliver Wang.
\newblock Neural scene flow fields for space-time view synthesis of dynamic scenes.
\newblock In \emph{CVPR}, pages 6498--6508, 2021.
\newblock arXiv:2011.13084.

\bibitem[Li et~al.(2023)Li, Wang, Cole, Tucker, and Snavely]{li2023dynibar}
Zhengqi Li, Qianqian Wang, Forrester Cole, Richard Tucker, and Noah Snavely.
\newblock {DynIBaR: Neural Dynamic Image-Based Rendering}.
\newblock In \emph{CVPR}, pages 4273--4284, 2023.

\bibitem[Li et~al.(2025)Li, Tucker, Cole, Wang, Jin, Ye, Kanazawa, Holynski, and Snavely]{li2025megasam}
Zhengqi Li, Richard Tucker, Forrester Cole, Qianqian Wang, Linyi Jin, Vickie Ye, Angjoo Kanazawa, Aleksander Holynski, and Noah Snavely.
\newblock {MegaSaM}: Accurate, fast, and robust structure and motion from casual dynamic videos.
\newblock In \emph{CVPR}, pages 10486--10496, 2025.

\bibitem[Lian et~al.(2023)Lian, Wu, and Yu]{lian2023rcf}
Long Lian, Zhirong Wu, and Stella~X. Yu.
\newblock Bootstrapping objectness from videos by relaxed common fate and visual grouping.
\newblock In \emph{CVPR}, pages 14582--14591, 2023.

\bibitem[Liang et~al.(2025{\natexlab{a}})Liang, Ren, Mirzaei, Torralba, Liu, Gilitschenski, Fidler, Oztireli, Ling, Gojcic, and Huang]{liang2025btimer}
Hanxue Liang, Jiawei Ren, Ashkan Mirzaei, Antonio Torralba, Ziwei Liu, Igor Gilitschenski, Sanja Fidler, Cengiz Oztireli, Huan Ling, Zan Gojcic, and Jiahui Huang.
\newblock Feed-forward bullet-time reconstruction of dynamic scenes from monocular videos.
\newblock In \emph{NeurIPS}, 2025{\natexlab{a}}.

\bibitem[Liang et~al.(2025{\natexlab{b}})Liang, Xu, and Kikuchi]{liang2025himor}
Yiming Liang, Tianhan Xu, and Yuta Kikuchi.
\newblock {HiMoR}: Monocular deformable {G}aussian reconstruction with hierarchical motion representation.
\newblock In \emph{CVPR}, pages 886--895, 2025{\natexlab{b}}.
\newblock arXiv:2504.06210.

\bibitem[Liang et~al.(2026)Liang, Zhu, Chang, Li, Lu, and Zhang]{liang2026reflow}
Yanzhe Liang, Ruijie Zhu, Hanzhi Chang, Zhuoyuan Li, Jiahao Lu, and Tianzhu Zhang.
\newblock {ReFlow}: Self-correction motion learning for dynamic scene reconstruction.
\newblock In \emph{CVPR}, pages 29803--29813, 2026.
\newblock arXiv:2604.01561.

\bibitem[Lin et~al.(2026)Lin, Lin, Pan, Yu, Hu, Yan, Fragkiadaki, and Mu]{lin2026movies}
Chenguo Lin, Yuchen Lin, Panwang Pan, Yifan Yu, Tao Hu, Honglei Yan, Katerina Fragkiadaki, and Yadong Mu.
\newblock {MoVieS}: Motion-aware {4D} dynamic view synthesis in one second.
\newblock In \emph{CVPR}, pages 295--306, 2026.

\bibitem[Liu et~al.(2026)Liu, Xiao, Chen, Feng, Tai, Tang, and Kang]{liu2026traceanything}
Xinhang Liu, Yuxi Xiao, Donny~Y. Chen, Jiashi Feng, Yu-Wing Tai, Chi-Keung Tang, and Bingyi Kang.
\newblock Trace anything: Representing any video in {4D} via trajectory fields.
\newblock In \emph{ICLR}, pages 37738--37767, 2026.
\newblock arXiv:2510.13802.

\bibitem[Lu et~al.(2026)Lu, Xu, Hu, Zhu, Zhao, Yeung, Shan, and Liu]{lu2026track4world}
Jiahao Lu, Jiayi Xu, Wenbo Hu, Ruijie Zhu, Chengfeng Zhao, Sai-Kit Yeung, Ying Shan, and Yuan Liu.
\newblock {Track4World}: Feedforward world-centric dense {3D} tracking of all pixels.
\newblock In \emph{ECCV}, pages 184--204, 2026.
\newblock arXiv:2603.02573.

\bibitem[Luiten et~al.(2024)Luiten, Kopanas, Leibe, and Ramanan]{luiten2024dynamic3dgs}
Jonathon Luiten, Georgios Kopanas, Bastian Leibe, and Deva Ramanan.
\newblock Dynamic {3D} {G}aussians: Tracking by persistent dynamic view synthesis.
\newblock In \emph{International Conference on 3D Vision (3DV)}, pages 800--809, 2024.

\bibitem[Luo et~al.(2026)Luo, Zhou, Lan, Pan, and Loy]{luo20264rc}
Yihang Luo, Shangchen Zhou, Yushi Lan, Xingang Pan, and Chen~Change Loy.
\newblock {4RC}: {4D} reconstruction via conditional querying anytime and anywhere.
\newblock In \emph{Int. Conf. Mach. Learn.}, pages 83198--83214, 2026.
\newblock arXiv:2602.10094.

\bibitem[Mazur et~al.(2026)Mazur, Taher, and Davison]{mazur2026primitivemache}
Kirill Mazur, Marwan Taher, and Andrew~J. Davison.
\newblock {4D} primitive-m{\^a}ch{\'e}: Glueing primitives for persistent {4D} scene reconstruction.
\newblock In \emph{CVPR}, pages 7372--7381, 2026.

\bibitem[Meunier et~al.(2023)Meunier, Badoual, and Bouthemy]{meunier2022emdriven}
Etienne Meunier, Ana{\"\i}s Badoual, and Patrick Bouthemy.
\newblock {EM}-driven unsupervised learning for efficient motion segmentation.
\newblock \emph{IEEE TPAMI}, 45\penalty0 (4):\penalty0 4462--4473, 2023.

\bibitem[Mildenhall et~al.(2020)Mildenhall, Srinivasan, Tancik, Barron, Ramamoorthi, and Ng]{mildenhall2020nerf}
Ben Mildenhall, Pratul~P. Srinivasan, Matthew Tancik, Jonathan~T. Barron, Ravi Ramamoorthi, and Ren Ng.
\newblock {NeRF}: Representing scenes as neural radiance fields for view synthesis.
\newblock In \emph{ECCV}, pages 405--421, 2020.

\bibitem[Nazarczuk et~al.(2025)Nazarczuk, Catley-Chandar, Tanay, Zhang, Slabaugh, and P{\'e}rez-Pellitero]{nazarczuk2025vidar}
Michal Nazarczuk, Sibi Catley-Chandar, Thomas Tanay, Zhensong Zhang, Gregory Slabaugh, and Eduardo P{\'e}rez-Pellitero.
\newblock {ViDAR}: Video diffusion-aware {4D} reconstruction from monocular inputs.
\newblock In \emph{NeurIPS}, pages 132090--132118, 2025.

\bibitem[Ngo et~al.(2025)Ngo, Zhuang, Kalogerakis, Gan, Tulyakov, Lee, and Wang]{ngo2024delta}
Tuan~Duc Ngo, Peiye Zhuang, Evangelos Kalogerakis, Chuang Gan, Sergey Tulyakov, Hsin-Ying Lee, and Chaoyang Wang.
\newblock {DELTA}: Dense efficient long-range {3D} tracking for any video.
\newblock In \emph{ICLR}, pages 44620--44641, 2025.

\bibitem[Ni et~al.(2026)Ni, Huang, Hou, Xie, and Zhang]{ni2026rayco4d}
Yuanxiang Ni, Xianliang Huang, Ruihan Hou, Feng Xie, and Hao Zhang.
\newblock {RayCo4D}: Ray-conditioned continuous motion model for monocular {4D} dynamic reconstruction.
\newblock \emph{arXiv preprint arXiv:2610.28351}, 2026.

\bibitem[Ochs et~al.(2014)Ochs, Malik, and Brox]{ochs2014segmentation}
Peter Ochs, Jitendra Malik, and Thomas Brox.
\newblock Segmentation of moving objects by long term video analysis.
\newblock \emph{IEEE TPAMI}, 36\penalty0 (6):\penalty0 1187--1200, 2014.

\bibitem[Park et~al.(2026)Park, Yun, Bui, and Kim]{park2026webspline}
Jongmin Park, Jeonghwan Yun, Minh-Quan~Viet Bui, and Munchurl Kim.
\newblock {WebSpline}: Structure-informed splines for real-time {3D} {G}aussians from monocular videos.
\newblock In \emph{NeurIPS}, 2026.
\newblock arXiv:2606.02096.

\bibitem[Park et~al.(2021{\natexlab{a}})Park, Sinha, Barron, Bouaziz, Goldman, Seitz, and Martin-Brualla]{park2021nerfies}
Keunhong Park, Utkarsh Sinha, Jonathan~T. Barron, Sofien Bouaziz, Dan~B Goldman, Steven~M. Seitz, and Ricardo Martin-Brualla.
\newblock Nerfies: Deformable neural radiance fields.
\newblock In \emph{ICCV}, pages 5865--5874, 2021{\natexlab{a}}.
\newblock arXiv:2011.12948.

\bibitem[Park et~al.(2021{\natexlab{b}})Park, Sinha, Hedman, Barron, Bouaziz, Goldman, Martin-Brualla, and Seitz]{park2021hypernerf}
Keunhong Park, Utkarsh Sinha, Peter Hedman, Jonathan~T. Barron, Sofien Bouaziz, Dan~B Goldman, Ricardo Martin-Brualla, and Steven~M. Seitz.
\newblock {HyperNeRF: A Higher-Dimensional Representation for Topologically Varying Neural Radiance Fields}.
\newblock \emph{ACM TOG}, 40\penalty0 (6):\penalty0 238:1--238:12, 2021{\natexlab{b}}.

\bibitem[Peng et~al.(2026)Peng, Lv, Lu, and Savva]{peng2026itaco}
Weikun Peng, Jun Lv, Cewu Lu, and Manolis Savva.
\newblock {iTACO}: Interactable digital twins of articulated objects from casually captured {RGBD} videos.
\newblock In \emph{International Conference on 3D Vision (3DV)}, pages 520--531, 2026.

\bibitem[Piccinelli et~al.(2024)Piccinelli, Yang, Sakaridis, Segu, Li, Van~Gool, and Yu]{piccinelli2024unidepth}
Luigi Piccinelli, Yung-Hsu Yang, Christos Sakaridis, Mattia Segu, Siyuan Li, Luc Van~Gool, and Fisher Yu.
\newblock {UniDepth}: Universal monocular metric depth estimation.
\newblock In \emph{CVPR}, pages 10106--10116, 2024.

\bibitem[Pumarola et~al.(2021)Pumarola, Corona, Pons-Moll, and Moreno-Noguer]{pumarola2021dnerf}
Albert Pumarola, Enric Corona, Gerard Pons-Moll, and Francesc Moreno-Noguer.
\newblock {D-NeRF}: Neural radiance fields for dynamic scenes.
\newblock In \emph{CVPR}, pages 10318--10327, 2021.

\bibitem[Qian et~al.(2026)Qian, Ren, Xiang, Ordonez, and Hang]{qian2026motionbits}
Howard~H. Qian, Kejia Ren, Yu Xiang, Vicente Ordonez, and Kaiyu Hang.
\newblock {MotionBits}: Video segmentation through motion-level analysis of rigid bodies.
\newblock \emph{arXiv preprint arXiv:2603.06846}, 2026.

\bibitem[Ravi et~al.(2025)Ravi, Gabeur, Hu, Hu, Ryali, Ma, Khedr, R{\"a}dle, Rolland, Gustafson, Mintun, Pan, Alwala, Carion, Wu, Girshick, Doll{\'a}r, and Feichtenhofer]{ravi2024sam2}
Nikhila Ravi, Valentin Gabeur, Yuan-Ting Hu, Ronghang Hu, Chaitanya Ryali, Tengyu Ma, Haitham Khedr, Roman R{\"a}dle, Chloe Rolland, Laura Gustafson, Eric Mintun, Junting Pan, Kalyan~Vasudev Alwala, Nicolas Carion, Chao-Yuan Wu, Ross Girshick, Piotr Doll{\'a}r, and Christoph Feichtenhofer.
\newblock {SAM} 2: Segment anything in images and videos.
\newblock In \emph{ICLR}, 2025.

\bibitem[Rozumny et~al.(2026)Rozumny, Luiten, Khan, Sch{\"o}nberger, and Kontschieder]{rozumny2026bulletgen}
Denis Rozumny, Jonathon Luiten, Numair Khan, Johannes Sch{\"o}nberger, and Peter Kontschieder.
\newblock {BulletGen}: Improving {4D} reconstruction with bullet-time generation.
\newblock In \emph{CVPRW}, pages 11617--11628, 2026.

\bibitem[Russell et~al.(2011)Russell, Fayad, and Agapito]{russell2011energy}
Chris Russell, Jo{\~a}o Fayad, and Lourdes Agapito.
\newblock Energy based multiple model fitting for non-rigid structure from motion.
\newblock In \emph{CVPR}, pages 3009--3016, 2011.

\bibitem[Russell et~al.(2014)Russell, Yu, and Agapito]{russell2014videopopup}
Chris Russell, Rui Yu, and Lourdes Agapito.
\newblock Video pop-up: Monocular {3D} reconstruction of dynamic scenes.
\newblock In \emph{ECCV}, pages 583--598, 2014.

\bibitem[Stearns et~al.(2024)Stearns, Harley, Uy, Dubost, Tombari, Wetzstein, and Guibas]{stearns2024marbles}
Colton Stearns, Adam Harley, Mikaela Uy, Florian Dubost, Federico Tombari, Gordon Wetzstein, and Leonidas Guibas.
\newblock {Dynamic Gaussian Marbles for Novel View Synthesis of Casual Monocular Videos}.
\newblock In \emph{SIGGRAPH Asia 2024 Conference Papers}, pages 1--11, 2024.

\bibitem[Strecke and St{\"u}ckler(2019)]{strecke2019emfusion}
Michael Strecke and J{\"o}rg St{\"u}ckler.
\newblock {EM-Fusion}: Dynamic object-level {SLAM} with probabilistic data association.
\newblock In \emph{ICCV}, pages 5864--5873, 2019.

\bibitem[Tomasi and Kanade(1992)]{tomasi1992shape}
Carlo Tomasi and Takeo Kanade.
\newblock Shape and motion from image streams under orthography: A factorization method.
\newblock \emph{IJCV}, 9\penalty0 (2):\penalty0 137--154, 1992.

\bibitem[Torr(1998)]{torr1998geometric}
Philip H.~S. Torr.
\newblock Geometric motion segmentation and model selection.
\newblock \emph{Philosophical Transactions of the Royal Society of London. Series A}, 356\penalty0 (1740):\penalty0 1321--1340, 1998.

\bibitem[Valmadre and Lucey(2012)]{valmadre2012general}
Jack Valmadre and Simon Lucey.
\newblock General trajectory prior for non-rigid reconstruction.
\newblock In \emph{CVPR}, pages 1394--1401, 2012.

\bibitem[Vidal et~al.(2005)Vidal, Ma, and Sastry]{vidal2005gpca}
Ren{\'e} Vidal, Yi Ma, and Shankar Sastry.
\newblock Generalized principal component analysis ({GPCA}).
\newblock \emph{IEEE TPAMI}, 27\penalty0 (12):\penalty0 1945--1959, 2005.

\bibitem[Wan et~al.(2024)Wan, Lu, and Zeng]{wan2024spgs}
Diwen Wan, Ruijie Lu, and Gang Zeng.
\newblock Superpoint {G}aussian splatting for real-time high-fidelity dynamic scene reconstruction.
\newblock In \emph{Int. Conf. Mach. Learn.}, pages 49957--49972, 2024.

\bibitem[Wang et~al.(2025{\natexlab{a}})Wang, Chen, Karaev, Vedaldi, Rupprecht, and Novotny]{wang2025vggt}
Jianyuan Wang, Minghao Chen, Nikita Karaev, Andrea Vedaldi, Christian Rupprecht, and David Novotny.
\newblock {VGGT}: Visual geometry grounded transformer.
\newblock In \emph{CVPR}, pages 5294--5306, 2025{\natexlab{a}}.

\bibitem[Wang et~al.(2026{\natexlab{a}})Wang, Chen, Zhang, Karaev, Sch{\"o}nberger, Labatut, Bojanowski, Novotny, Vedaldi, and Rupprecht]{wang2026vggtomega}
Jianyuan Wang, Minghao Chen, Shangzhan Zhang, Nikita Karaev, Johannes Sch{\"o}nberger, Patrick Labatut, Piotr Bojanowski, David Novotny, Andrea Vedaldi, and Christian Rupprecht.
\newblock {VGGT}-$\omega$.
\newblock In \emph{CVPR}, pages 21486--21499, 2026{\natexlab{a}}.

\bibitem[Wang and Adelson(1994)]{wang1994layers}
John Y.~A. Wang and Edward~H. Adelson.
\newblock Representing moving images with layers.
\newblock \emph{IEEE TIP}, 3\penalty0 (5):\penalty0 625--638, 1994.

\bibitem[Wang et~al.(2025{\natexlab{b}})Wang, Ye, Gao, Zeng, Austin, Li, and Kanazawa]{wang2024shapeofmotion}
Qianqian Wang, Vickie Ye, Hang Gao, Weijia Zeng, Jake Austin, Zhengqi Li, and Angjoo Kanazawa.
\newblock {Shape of Motion: 4D Reconstruction from a Single Video}.
\newblock In \emph{ICCV}, pages 9660--9672, 2025{\natexlab{b}}.
\newblock arXiv:2407.13764.

\bibitem[Wang et~al.(2025{\natexlab{c}})Wang, Zhang, Holynski, Efros, and Kanazawa]{wang2025cut3r}
Qianqian Wang, Yifei Zhang, Aleksander Holynski, Alexei~A. Efros, and Angjoo Kanazawa.
\newblock Continuous {3D} perception model with persistent state.
\newblock In \emph{CVPR}, pages 10510--10522, 2025{\natexlab{c}}.

\bibitem[Wang et~al.(2026{\natexlab{b}})Wang, Zhao, and Li]{wang2026worldtree}
Qisen Wang, Yifan Zhao, and Jia Li.
\newblock {WorldTree}: Towards {4D} dynamic worlds from monocular video using tree-chains.
\newblock In \emph{ICLR}, 2026{\natexlab{b}}.
\newblock arXiv:2602.11845.

\bibitem[Weiss and Adelson(1996)]{weiss1996unified}
Yair Weiss and Edward~H. Adelson.
\newblock A unified mixture framework for motion segmentation: Incorporating spatial coherence and estimating the number of models.
\newblock In \emph{CVPR}, pages 321--326, 1996.

\bibitem[Wu et~al.(2026{\natexlab{a}})Wu, Li, Chen, and Pfister]{wu2026rigs}
Chenyu Wu, Wanhua Li, Zhu-Tian Chen, and Hanspeter Pfister.
\newblock {RiGS}: Rigid-aware {4D} {G}aussian splatting from a single monocular video.
\newblock In \emph{Proceedings of the IEEE/CVF Conference on Computer Vision and Pattern Recognition (CVPR) Findings}, pages 547--557, 2026{\natexlab{a}}.
\newblock arXiv:2605.23672.

\bibitem[Wu et~al.(2025{\natexlab{a}})Wu, Liu, Hung, Qian, Zhan, and Duan]{wu2025fourdfly}
Diankun Wu, Fangfu Liu, Yi-Hsin Hung, Yue Qian, Xiaohang Zhan, and Yueqi Duan.
\newblock {4D-Fly}: Fast {4D} reconstruction from a single monocular video.
\newblock In \emph{CVPR}, pages 16663--16673, 2025{\natexlab{a}}.

\bibitem[Wu et~al.(2024)Wu, Yi, Fang, Xie, Zhang, Wei, Liu, Tian, and Wang]{wu2024fourdgs}
Guanjun Wu, Taoran Yi, Jiemin Fang, Lingxi Xie, Xiaopeng Zhang, Wei Wei, Wenyu Liu, Qi Tian, and Xinggang Wang.
\newblock {4D} {G}aussian splatting for real-time dynamic scene rendering.
\newblock In \emph{CVPR}, pages 20310--20320, 2024.
\newblock arXiv:2310.08528.

\bibitem[Wu et~al.(2025{\natexlab{b}})Wu, Tao, Wang, Liu, Kompella, and Yan]{wu2025origs}
Junyi Wu, Jiachen Tao, Haoxuan Wang, Gaowen Liu, Ramana~Rao Kompella, and Yan Yan.
\newblock Orientation-anchored hyper-{G}aussian for {4D} reconstruction from casual videos.
\newblock In \emph{NeurIPS}, pages 144115--144138, 2025{\natexlab{b}}.
\newblock arXiv:2509.23492.

\bibitem[Wu et~al.(2025{\natexlab{c}})Wu, Gao, Poole, Trevithick, Zheng, Barron, and Holynski]{wu2025cat4d}
Rundi Wu, Ruiqi Gao, Ben Poole, Alex Trevithick, Changxi Zheng, Jonathan~T. Barron, and Aleksander Holynski.
\newblock {CAT4D}: Create anything in {4D} with multi-view video diffusion models.
\newblock In \emph{CVPR}, pages 26057--26068, 2025{\natexlab{c}}.

\bibitem[Wu et~al.(2026{\natexlab{b}})Wu, Yan, Yi, Wang, and Liao]{wu2026streamsplat}
Zike Wu, Qi Yan, Xuanyu Yi, Lele Wang, and Renjie Liao.
\newblock {StreamSplat}: Towards online dynamic {3D} reconstruction from uncalibrated video streams.
\newblock In \emph{ICLR}, 2026{\natexlab{b}}.
\newblock arXiv:2506.08862.

\bibitem[Xiao et~al.(2024)Xiao, Wang, Zhang, Xue, Peng, Shen, and Zhou]{xiao2024spatialtracker}
Yuxi Xiao, Qianqian Wang, Shangzhan Zhang, Nan Xue, Sida Peng, Yujun Shen, and Xiaowei Zhou.
\newblock {SpatialTracker}: Tracking any {2D} pixels in {3D} space.
\newblock In \emph{CVPR}, pages 20406--20417, 2024.

\bibitem[Xie et~al.(2022)Xie, Xie, and Zisserman]{xie2022oclr}
Junyu Xie, Weidi Xie, and Andrew Zisserman.
\newblock Segmenting moving objects via an object-centric layered representation.
\newblock In \emph{NeurIPS}, pages 28023--28036, 2022.

\bibitem[Xie et~al.(2024)Xie, Yang, Xie, and Zisserman]{xie2024flowsam}
Junyu Xie, Charig Yang, Weidi Xie, and Andrew Zisserman.
\newblock Moving object segmentation: All you need is {SAM} (and flow).
\newblock In \emph{ACCV}, pages 162--178, 2024.

\bibitem[Xu et~al.(2025)Xu, Li, Dong, Zhou, Newcombe, and Lv]{xu2025fourdgt}
Zhen Xu, Zhengqin Li, Zhao Dong, Xiaowei Zhou, Richard Newcombe, and Zhaoyang Lv.
\newblock {4DGT}: Learning a {4D} {G}aussian transformer using real-world monocular videos.
\newblock In \emph{NeurIPS}, 2025.

\bibitem[Yang et~al.(2021)Yang, Lamdouar, Lu, Zisserman, and Xie]{yang2021motiongrouping}
Charig Yang, Hala Lamdouar, Erika Lu, Andrew Zisserman, and Weidi Xie.
\newblock Self-supervised video object segmentation by motion grouping.
\newblock In \emph{ICCV}, pages 7177--7188, 2021.

\bibitem[Yang et~al.(2023)Yang, Gao, Li, Gao, Wang, and Zheng]{yang2023trackanything}
Jinyu Yang, Mingqi Gao, Zhe Li, Shang Gao, Fangjing Wang, and Feng Zheng.
\newblock Track anything: Segment anything meets videos.
\newblock \emph{arXiv preprint arXiv:2304.11968}, 2023.

\bibitem[Yang et~al.(2024{\natexlab{a}})Yang, Zhang, Huang, Chen, Xu, and Su]{yang2024movingparts}
Kaizhi Yang, Xiaoshuai Zhang, Zhiao Huang, Xuejin Chen, Zexiang Xu, and Hao Su.
\newblock {MovingParts}: Motion-based {3D} part discovery in dynamic radiance field.
\newblock In \emph{ICLR}, 2024{\natexlab{a}}.

\bibitem[Yang et~al.(2024{\natexlab{b}})Yang, Kang, Huang, Xu, Feng, and Zhao]{yang2024depthanything}
Lihe Yang, Bingyi Kang, Zilong Huang, Xiaogang Xu, Jiashi Feng, and Hengshuang Zhao.
\newblock {Depth Anything: Unleashing the Power of Large-Scale Unlabeled Data}.
\newblock In \emph{CVPR}, pages 10371--10381, 2024{\natexlab{b}}.

\bibitem[Yang et~al.(2024{\natexlab{c}})Yang, Gao, Zhou, Jiao, Zhang, and Jin]{yang2024deformable3dgs}
Ziyi Yang, Xinyu Gao, Wen Zhou, Shaohui Jiao, Yuqing Zhang, and Xiaogang Jin.
\newblock {Deformable 3D Gaussians for High-Fidelity Monocular Dynamic Scene Reconstruction}.
\newblock In \emph{CVPR}, pages 20331--20341, 2024{\natexlab{c}}.

\bibitem[Yoon et~al.(2020)Yoon, Kim, Gallo, Park, and Kautz]{yoon2020novel}
Jae~Shin Yoon, Kihwan Kim, Orazio Gallo, Hyun~Soo Park, and Jan Kautz.
\newblock {Novel View Synthesis of Dynamic Scenes with Globally Coherent Depths from a Monocular Camera}.
\newblock In \emph{CVPR}, pages 5335--5344, 2020.

\bibitem[Yu et~al.(2026)Yu, Yang, Luo, Sun, and Hariharan]{yu2026smg}
Haozheng Yu, Xinyu Yang, Rundong Luo, Jennifer~J. Sun, and Bharath Hariharan.
\newblock {SMG}: Semantic motion graph for monocular dynamic {G}aussian splatting.
\newblock In \emph{ECCV}, pages 151--171, 2026.
\newblock arXiv:2608.31023.

\bibitem[Yu et~al.(2025)Yu, Hu, Xing, and Shan]{yu2025trajectorycrafter}
Mark Yu, Wenbo Hu, Jinbo Xing, and Ying Shan.
\newblock {TrajectoryCrafter}: Redirecting camera trajectory for monocular videos via diffusion models.
\newblock In \emph{ICCV}, pages 100--111, 2025.
\newblock arXiv:2503.05638.

\bibitem[Zhang et~al.(2026{\natexlab{a}})Zhang, Le~Moing, Koppula, Rocco, Momeni, Xie, Sun, Sukthankar, Barral, Hadsell, Ghahramani, Zisserman, Zhang, and Sajjadi]{zhang2026d4rt}
Chuhan Zhang, Guillaume Le~Moing, Skanda Koppula, Ignacio Rocco, Liliane Momeni, Junyu Xie, Shuyang Sun, Rahul Sukthankar, Jo{\"e}lle~K. Barral, Raia Hadsell, Zoubin Ghahramani, Andrew Zisserman, Junlin Zhang, and Mehdi S.~M. Sajjadi.
\newblock Efficiently reconstructing dynamic scenes one {D4RT} at a time.
\newblock In \emph{CVPR}, pages 7382--7392, 2026{\natexlab{a}}.

\bibitem[Zhang et~al.(2025{\natexlab{a}})Zhang, Herrmann, Hur, Jampani, Darrell, Cole, Sun, and Yang]{zhang2025monst3r}
Junyi Zhang, Charles Herrmann, Junhwa Hur, Varun Jampani, Trevor Darrell, Forrester Cole, Deqing Sun, and Ming-Hsuan Yang.
\newblock {MonST3R}: A simple approach for estimating geometry in the presence of motion.
\newblock In \emph{ICLR}, 2025{\natexlab{a}}.

\bibitem[Zhang et~al.(2025{\natexlab{b}})Zhang, Chang, Liu, and Boularias]{zhang2025motionblender}
Xinyu Zhang, Haonan Chang, Yuhan Liu, and Abdeslam Boularias.
\newblock {Motion Blender} {G}aussian splatting for dynamic reconstruction.
\newblock In \emph{Conference on Robot Learning (CoRL)}, pages 5107--5128, 2025{\natexlab{b}}.
\newblock arXiv:2503.09040.

\bibitem[Zhang et~al.(2026{\natexlab{b}})Zhang, Xiao, Huang, Zheng, and Zhang]{zhang2026se3bsplinegs}
Xuankai Zhang, Junjin Xiao, Shangwei Huang, Wei-shi Zheng, and Qing Zhang.
\newblock Learning explicit continuous motion representation for dynamic {G}aussian splatting from monocular videos.
\newblock In \emph{CVPR}, pages 33291--33300, 2026{\natexlab{b}}.
\newblock arXiv:2603.25058.

\bibitem[Zhao et~al.(2026)Zhao, Deng, Wei, and Lin]{zhao2026ground4d}
Qing Zhao, Weijian Deng, Pengxu Wei, and Liang Lin.
\newblock {Ground4D}: Consistency-aware {4D} reconstruction from monocular video.
\newblock \emph{arXiv preprint arXiv:2606.28828}, 2026.

\bibitem[Zhou and Lee(2025)]{zhou2025motion4d}
Haoran Zhou and Gim~Hee Lee.
\newblock {Motion4D}: Learning {3D}-consistent motion and semantics for {4D} scene understanding.
\newblock In \emph{NeurIPS}, pages 116032--116051, 2025.
\newblock arXiv:2512.03601.

\bibitem[Zhou and Lee(2026)]{zhou2026motionscale}
Haoran Zhou and Gim~Hee Lee.
\newblock {MotionScale}: Reconstructing appearance, geometry, and motion of dynamic scenes with scalable {4D} {G}aussian splatting.
\newblock In \emph{CVPR}, pages 11860--11870, 2026.
\newblock arXiv:2603.29296.

\bibitem[Zhu et~al.(2026)Zhu, Huang, Mazumdar, Li, Gojcic, Wetzstein, Armeni, De~Mello, and Trevithick]{zhu2026worldfrommotion}
Liyuan Zhu, Shengyu Huang, Amrita Mazumdar, Tianye Li, Zan Gojcic, Gordon Wetzstein, Iro Armeni, Shalini De~Mello, and Alex Trevithick.
\newblock World from motion: Generative dynamic {G}aussian reconstruction from monocular video.
\newblock In \emph{NeurIPS}, 2026.

\bibitem[Zhu and Lucey(2015)]{zhu2015convolutional}
Yingying Zhu and Simon Lucey.
\newblock Convolutional sparse coding for trajectory reconstruction.
\newblock \emph{IEEE TPAMI}, 37:\penalty0 529--540, 2015.

\bibitem[Zhuo et~al.(2026)Zhuo, Zheng, Guo, Wu, Zhou, and Lu]{zhuo2026streamvggt}
Dong Zhuo, Wenzhao Zheng, Jiahe Guo, Yuqi Wu, Jie Zhou, and Jiwen Lu.
\newblock Streaming visual geometry transformer.
\newblock In \emph{ICLR}, 2026.

\end{thebibliography}
}

\end{document}